# Deep Learning for Satellite Image Time Series Analysis: A Review

Lynn Miller, *Member, IEEE*, Charlotte Pelletier, and Geoffrey I. Webb, *Fellow, IEEE*

Earth observation (EO) satellite missions have been providing detailed images about the state of the Earth and its land cover for over 50 years. Long term missions, such as NASA's Landsat, Terra, and Aqua satellites, and more recently, the ESA's Sentinel missions, record images of the entire world every few days. Although single images provide point-in-time data, repeated images of the same area, or satellite image time series (SITS) provide information about the changing state of vegetation and land use. These SITS are useful for modeling dynamic processes and seasonal changes such as plant phenology. They have potential benefits for many aspects of land and natural resource management, including applications in agricultural, forest, water, and disaster management, urban planning, and mining. However, the resulting satellite image time series (SITS) are complex, incorporating information from the temporal, spatial, and spectral dimensions. Therefore, deep learning methods are often deployed as they can analyze these complex relationships. This review presents a summary of the state-of-the-art methods of modelling environmental, agricultural, and other Earth observation variables from SITS data using deep learning methods. We aim to provide a resource for remote sensing experts interested in using deep learning techniques to enhance Earth observation models with temporal information.

In this review, we are primarily concerned with methods of estimating or predicting EO variables from SITS data. These can be divided into two types of tasks depending on the nature of the variable being estimated. If the variable can take one of two or more discrete values, then the task is classification. Examples of classification tasks include land cover mapping [1], crop type identification [2], and burnt area mapping [3]. If the variable can take continuous numeric values, then the task is regression. In the context of time series, regression tasks can be further categorized as extrinsic regression tasks, which estimate the value of a variable external to those represented by the time series [4], or forecasting tasks, which predict future values of a time series based on its historical values. While classification and extrinsic regression tasks are technically distinct, in practice many of the deep learning methods used are very similar. Many architectures that have originally been designed for a classification task can easily be adapted for extrinsic regression tasks (and vice versa) [5], for example, by modifying the last layer and the loss function.

A more important consideration when considering deep learning architectures for SITS tasks is the quantity of labeled data available for training models. Many deep learning models have thousands or even millions of parameters that need estimating and thus training these models require large quantities of labeled data. Smaller architectures with fewer parameters are likely to be more suitable when labeled data are limited. In particular, techniques such as semi-supervised and unsupervised learning are designed for situations with few or no labeled samples, respectively.

In related work, Gómez et al. [1] provided a comprehensive review of using optical SITS data for land cover classification. However, there have been developments in EO data and machine learning since that review that have led to a substantial increase in SITS research and its potential applications for EO monitoring. One reason for these recent developments is the availability of data from the ESA Sentinel missions that provide both optical and synthetic aperture radar (SAR) data at higher temporal and spatial resolution than many of the previously readily available sources. Another reason is the wide variety of machine learning methods, especially deep learning methods, that can model the complex relationships that exist between the observed electromagnetic radiation and the variable of interest. Both these advances mean there are a wider variety of techniques available for EO modelling and a wider variety of tasks that can be performed using these models.

The current review, which covers the use of deep learning methods for SITS, therefore provides an update to review [1]. It focuses on deep learning analysis of SITS for classification and extrinsic regression problems and examines a broad range of applications of SITS data, thus filling the gap left by these other recent reviews. However, the review excludes DL forecasting applications of SITS, as these have been extensively covered by Moskolaï et al. [6]. As we are interested in modelling of temporal features, we limit the study to time series longer than two. Thus, we exclude methods such as bitemporal change detection, which identifies differences in two images obtained at separate times. A recent review of change detection in remote sensing is provided in [7].

## SELECTION PRINCIPLES AND RELATED SURVEYS

### SELECTION PRINCIPLES

There are more studies using deep learning for SITS than can feasibly be included in a single review, thus this review is not an exhaustive survey. However, we aim to provide coverage of a broad range of studies that show both the deep learning methods applied to SITS and the tasks for which SITS have been used. We have therefore included studies that: 1) are the key works developing DL techniques for SITS tasks, 2) show how the various DL methods have been applied to SITS, 3) provide insight into extracting temporal and/or spatial features from SITS, and 4) highlight the wide range of tasks for which SITS can be used. Papers were mainly found by searching on Google Scholar, other resources used were Web of Science and Scopus, plus the authors' knowledge of relevant studies, drawing on their prior knowledge of the subject.

*RELATED SURVEYS*

Previous works have reviewed machine learning and deep learning methods for modelling remote sensing images. However, none of these reviews fully cover the scope of this review – deep learning from SITS with a focus on classification and extrinsic regression tasks (Table 1). Gomez-Chova et al. [8] review remote sensing classification using multiple heterogeneous image sources. The techniques reviewed are increasingly relevant as more remote sensing image sources become available. Gómez et al. [1] is an older review highlighting the importance of SITS data and machine learning methods for land cover classification. Our current review is, in many ways, a follow-up to the Gómez et al. review, with a focus on deep learning techniques, however we also review the use of SITS in a broader range of applications. Zhu et al. [9] reviewed the advances and challenges in DL for remote sensing, and the resources available that are potentially useful to help DL address some of the major challenges facing humanity. Li et al. reviewed deep learning techniques for pixel-wise and scene-wise image classification [10]. Ma et al. [11] studied the role of deep learning in Earth observation using remotely sensed data. It covers a broad range of tasks, including image fusion, image segmentation and object-based analysis, as well as classification tasks.

Table 1: Related Earth observation reviews and surveys.

| Title | Year and Reference | SITS | DL | Classification | Extrinsic Regression |
|---|---|---|---|---|---|
| Multimodal Classification of Remote Sensing Images: A Review and Future Directions | 2015 [8] | Partial | Partial | Partial | No |
| Optical remotely sensed time series data for land cover classification: A review | 2016 [1] | Yes | No | Partial | No |
| Deep learning in remote sensing: A comprehensive review and list of resources & 2017 | 2017 [9] | Partial | Yes | No | No |
| Deep learning for remote sensing image classification: A survey | 2018 [10] | No | Yes | No | No |
| Deep learning in remote sensing applications: A meta-analysis and review | 2019 [11] | Partial | Yes | Partial | No |
| Deep learning in environmental remote sensing: Achievements and challenges | 2020 [12] | Partial | Yes | No | No |
| Knowledge discovery from remote sensing images: A review | 2020 [13] | Partial | Partial | No | No |
| Recent Applications of Landsat 8/OLI and Sentinel-2/MSI for Land Use and Land Cover Mapping: A Systematic Review | 2020 [14] | Yes | No | Partial | Partial |
| Application of Deep Learning Architectures for Satellite Image Time Series Prediction: A Review | 2021 [6] | Yes | Yes | No | No |
| Review on Convolutional Neural Networks (CNN) in vegetation remote sensing | 2021 [15] | Partial | Partial | No | No |
| Artificial Intelligence for Remote Sensing Data Analysis: A review of challenges and opportunities | 2022 [16] | Partial | Yes | No | No |
| Deep Learning and Earth Observation to Support the Sustainable Development Goals: Current approaches, open challenges, and future opportunities | 2022 [17] | Partial | Yes | Partial | Partial |
| Deep learning for processing and analysis of remote sensing big data: a technical review | 2022 [18] | Partial | Yes | Partial | No |

In more recent reviews, Yuan et al. [12] reviewed DL applications for remote sensing, comparing the role of DL versus physical modelling of environmental variables and highlighting challenges in DL for remote sensing that need to be addressed. Wang et al. [13] provided a broad review of knowledge discovery techniques for remote sensing images, ranging from rules-based algorithms through to deep learning and ensembling techniques. Chaves et al. [14] reviewed recent research using Landsat 8 and/or Sentinel-2 data for land cover mapping. While not focused on SITS DL methods, the review noted the growing importance of these methods. Moskolaï et al. [6] is a review of forecasting applications using DL with SITS data. It provided an analysis of the main DL architectures, many of which are relevant for classification as well as forecasting.

Kattenborn et al. [15] provided an in-depth review of the use of convolutional neural networks in remote sensing for vegetation-related applications. Zhang and Zhang [16] reviewed the use of artificial intelligence (AI) in remote sensing data analysis, covering a range of topics including computational intelligence, natural language processing, and AI explicability and security, in addition to machine learning. Persello et al. [17] discussed the potential of deep learning and Earth observation to help address some of the major challenges facing humanity, particularly their role in supporting the UN sustainable development goals. Zhang et al. [18] is a recent review of deep learning techniques for remote sensing, with an emphasis on the processing of big data. It provided some coverage of SITS but did not explore the full range of SITS techniques and applications.

SATELLITE IMAGE TIMES SERIES

*SATELLITE EARTH OBSERVATION*

EO data is collected by instruments carried on over 1200 satellites [19] that observe the Earth at different spatial and temporal resolutions, and spectral frequencies [20] and measure a diverse range of geophysical parameters [21], complementing in-situ and other traditional methods of environmental monitoring. Although many of these satellites are commercial, many others are operated by national space agencies, some of which make their data freely available. This freely available data is of great importance as it allows modelling of environmental indicators at many scales, from local to global, thus providing benefits to countries and other organizations unable to afford the high cost associated with much commercial imagery [21].

Most satellites used for Earth observation follow a low-earth orbit [22], at about 600-1000km and orbit the Earth once every 90-100 minutes. They follow a polar or near-polar orbit, and so provide coverage of almost all the Earth's surface, and many are sun-synchronous, meaning they cross the equator at the same local time on each orbit. Each satellite will take several days to provide full coverage of the Earth, however multiple satellites may work together as a constellation to provide more frequent coverage [14]. There is usually a trade-off for sensors between spatial-spectral resolutions and high frequency of acquisitions over large areas. In addition to constraints related to satellite orbits and recording capacities, sensors are subject to several trade-offs between spectral resolution and signal-to-noise ratio, between spatial resolution and the volume of data to be stored, and between spatial resolution and temporal resolution.

The first sensors capable of acquiring time series of satellite images over large areas were mainly dedicated to meteorology. For example, the Advanced Very High Resolution Radiometer (AVHRR) sensor, which equipped the National Oceanic and Atmospheric Administration (NOAA) satellites since the 1970s, can acquire images with kilometric resolution over a very wide field of observation (around 3,000 kilometers) in six wavelengths [23]. All surfaces are imaged at least once a day.

France, in collaboration with Belgium and Sweden, also launched an Earth observation program in 1978: Satellites for Earth Observation (SPOT). The images from the first three satellites included three spectral bands with a resolution of twenty meters, and covered areas of 3600 km². The SPOT-4 and -5 satellites provided images for four spectral bands at 10- and 20-meters resolution, respectively. SPOT data was used in the Assimilation of Spatial Data within Agriculture Models (ADAM) experiment [24], [25]. This experiment was one of the first to investigate the capabilities of high spatial and temporal resolution data for modelling vegetation phenology. The experiment showed the changes in optical response, particularly in the NIR and Red bands, throughout the wheat growing season [24] and the potential for this to be exploited to model vegetation variables such as leaf area index [25]. During their end-of-life missions, the two SPOT-4 and -5 satellites were slightly lowered in orbit to enable acquisitions to be made every five days over more than a hundred sites: the Take-5 experiment [26]. These acquisitions provided datasets useful for developing the methods and techniques needed to process data from future satellite missions, notably Sentinel-2.

During the 2000s, the spatial resolution of sensors with high temporal revisit reached 250 to 300 meters, as in the case of the Moderate Resolution Imaging Spectroradiometer (MODIS) sensors on the American satellites Terra and Aqua, and the Medium Resolution Imaging Spectrometer (MERIS) on the European satellite ENVIronment SATellite (ENVISAT). More specifically, the MODIS sensor provides a daily image of all land surfaces in thirty-six spectral bands at resolutions ranging from 250 meters to one kilometer. The high repeatability of acquisitions and the global coverage of these images make MERIS and MODIS the sensors of choice for studying land surfaces [27]. However, the low spatial resolution of these sensors prevents their use for some land monitoring applications [28].

At the same time, very high spatial resolution sensors - SPOT-6-7, Pleiades, Quickbird, Ikonos and WorldView - and very high spectral resolution (known as hyper-spectral sensors) have also been developed. Thanks to their agility, very high spatial resolution sensors are particularly well suited to three-dimensional mapping of urban areas, monitoring sensitive sites or areas vulnerable to geophysical hazards. However, the images that form the SITS will be taken with different viewing angles, complexifying their analysis. Additionally, the small swaths and the long revisit time of these sensors prevent their use to monitor landscape dynamics over large areas.

Meanwhile, projects dedicated to Earth observation using high spatial resolution and medium temporal resolution satellites were developed to provide frequent observations of land surfaces. The first was the civilian Landsat program launched by the National Aeronautics and Space Administration (NASA) in 1972 [29]. Initially dedicated to assessing cereal crops in the United States and the former USSR (Union of Soviet Socialist Republics), this program can now be used to study all continental surfaces. A total of nine satellites were launched between 1972 and 2021, three of which are still in orbit – Landsat-7, -8 and -9. The Operational Land Imager (OLI) sensors on Landsat-8 and -9 provide images with a resolution of thirty meters in nine spectral bands, with a revisit time of sixteen days [30].

Since 2008, the European Space Agency (ESA) has been responsible for development and delivery of Sentinel satellites to meet part of the needs of the European Earth monitoring program Copernicus [31], [32]. More specifically, recent acquisitions by the Sentinel-2 satellites include thirteen spectral bands in the visible and infrared at a resolution of 10 to 60 meters over a swath of 290 kilometers. The novelty provided by Sentinel-2 lies in the combination of these high spectral and spatial resolutions with high temporal revisit - the two Sentinel-2 satellites cover the entire land area every five days [33].

Table 2. The main Earth Observation satellites and instruments used by studies reviewed by this survey.

| Agency | Satellite | Instrument[1,2] | Type[1] | Spatial Resolution[1] | Revisit Time[1] | Spectral bands[1] |
|---|---|---|---|---|---|---|
| NASA | Landsat-5 [34] | Thematic Mapper | Optical | 30m | 16 days | 7 VIS/IR |
| | Landsat-7 [34] | Enhanced Thematic Mapper Plus | Optical | 30m + 15m PAN | 16 days | 8 VIS/IR, 1 PAN |
| | Landsat-8 [34] | Operational Land Imager | Optical | 30m + 15m PAN | 16 days | 9 VIS/IR, 1 PAN |
| | Landsat-9 [34] | Operational Land Imager-2 | Optical | 30m + 15m PAN | 16 days | 9 VIS/IR, 1 PAN |
| | NOAA-7, 9, 11—14 | AVHRR-2 [35] | Optical | 1.1km | 1 day | 5 VIS/IR |
| | Terra/Aqua | MODIS [36] | Optical | 250m, 500m & 1000m | 1 day | 8 VIS/IR |
| | Suomi NPP | VIIRS [37] | Optical | 400m & 1.6km | 1 day | 22 VIS/IR |
| ESA | Sentinel-1A & 1B [38] | C-band synthetic-aperture radar | Microwave SAR | 10m | 6 days | Microwave C-band |
| | Sentinel-2A & 2B [38] | MultiSpectral Instrument | Optical | 10m, 20m & 60m | 5 days | 13 VIS/IR |
| CNES (France) | SPOT 4 [39] | Visible & Infrared High-Resolution | Optical | 20m + 10m PAN | 5 days[3] | 4 VIS/IR, 1 PAN |
| | SPOT 5 [39] | High-Resolution Geometric | Optical | 10m + 5m PAN | 5 days[3] | 4 VIS/IR, 1 PAN |
| CNES (France) & ISA (Israel) | Venµs [40] | Multispectral camera | Optical | 5-10m | 2 days | 12 VIS/IR |
| DLR (Germany) | TerraSAR-X [41] TanDEM-X [42] | X-band radar sensor | Microwave SAR | 1-2m | 11 days | Microwave X-band |
| NSPO (Taiwan) | Formosat-2 [43] | Remote Sensing Instrument | Optical | 8m + 2m PAN | 1 day | 4 VIS/IR, 1 PAN |
| | Formosat-5 [44] | Remote Sensing Instrument | Optical | 4m + 2m PAN | 2 days | 4 VIS/IR, 1 PAN |
| CRESDA (China) | Gaofen 1 [45], [46] | Panchromatic and Multi-spectral CCD Camera | Optical | 8m + 2m PAN | 1 month | 4 VIS/IR, 1 PAN |
| | Gaofen 2 [45], [46] | Panchromatic and Multi-spectral CCD Camera 2 | Optical | 3m + 0.75m PAN | 2 months | 4 VIS/IR, 1 PAN |
| CRESDA (China) & INPE (Brazil) | CBERS-4 & 4A [47] | Multispectral Camera | Optical | 20m | 26 days | 4 VIS/IR |
| | | Wide Field Imager-2 | Optical | 73m | 5 days | 4 VIS/IR |
| JAXA (Japan) | ALOS-2 [48] | PALSAR-2 | Microwave SAR | 3-10m | 14 days | Microwave L-band |
| Planet | 200 Flocks [49] | Flock | Optical | 3.7m | 1 day | 4 VIS/IR |

1. Details obtained from references cited in the table and the Committee on Earth Observation Satellites (CEOS) database [50].
2. Many satellites carry multiple instruments. The table lists only the ones used in the reviewed studies.
3. After implementation of the Take-5 experiment [26].

One method of resolving the trade-off between spatial and temporal resolutions is to deploy several satellites equipped with the same sensors and operating at regularly spaced orbits, known as a constellation. Both the Landsat and Sentinel-2 missions operate as small constellations. A commercial space agency, Planet, operates a constellation of about two hundred satellites, providing daily images in four spectral bands at 3.7m resolution.

While the above-mentioned sensors allow acquiring optical images (in visible-light and higher-infrared (IR) frequencies), microwave remote sensing instruments also allow the acquisition of dense time series. Notable examples are the ESA's Sentinel-1/SAR-C [32] and the Japan Aerospace Exploration Agency's (JAXA) ALOS/PALSAR [51], both synthetic aperture radar (SAR) instruments, which provide high resolution images every six and 14 days, respectively. Additionally, current developments in sensors will allow the acquisition of time series of hyper-spectral images or 3D point clouds. Examples include the ESA's Sentinel-3 Ocean and Land Colour Instrument (OCLI) [52] and NASA's ICESat-2 (Ice, Cloud, and land Elevation Satellite-2) Advanced Topographic Laser Altimeter System (ATLAS) [53]. These will provide time series of images suitable for ocean and cryosphere applications, in addition to land surface tasks.

Table 2 lists the main satellites and instruments used in the studies reviewed for this survey, together with details of their spatial, temporal, and spectral resolutions.

*SITS RESOURCES*

Earth Observation Platforms

Raw data collected by satellite instruments needs to be pre-processed before being used in machine learning. This is frequently done by the data providers to produce analysis ready datasets (ARD). Common preprocessing steps include [54]: spatial alignment or georeferencing; radiance correction (for example, to remove sun glare); quality control – e.g., adding quality flags. Further processing, such as that done by country or regional Data Cubes [54]–[58], include co-registration of images (spatially aligning them and converting them to the same resolution and projection). This enables data users to easily extract data for required locations and times from multiple data sources in a consistent manner, facilitating both multi-temporal and multi-modal data analysis without the need for complex pre-processing.

Many countries and regions have instigated data cube initiatives. These make analysis ready EO data from a range of satellite instruments for specific regions available in a consistent format. They allow access to a range of datasets that can be used to investigate issues of local importance [59], provide EO products tailored for local requirements [60], while allowing countries to maintain control over their data and meet regulatory and reporting requirements [61]. While many of these make use of open data cube [62], which provides open software tools for EO data [57], some challenges exist to provide interoperability between local data cubes [60].

Earth observation datasets are large, especially when high-resolution data for large regions are used. Processing these datasets can quickly exceed the capacity of personal computers and workstations [63]. Cloud-based solutions provide access to the large storage and computing resources needed for analyzing these large datasets. These platforms include large cloud computing providers such as Earth on Amazon Web Service (AWS) [64], Microsoft's Planetary Computer [65], and Google Earth Engine (GEE) [66], as well as special-purpose cloud providers such as Sentinel-Hub [67]. Further information about these platforms is provided in [63] and [68], including a summary of the capabilities, strengths, and weaknesses of each one.

SITS Datasets and Benchmarks

Studies applying machine learning to SITS have used datasets from many different regions, times, and satellite sensors. While these studies have produced important results for specific applications, the differences in the study regions make it difficult to compare the advantages and disadvantages of the various algorithms. Freely available benchmark datasets provide a means of evaluating and comparing algorithms.

Following the trend in remote sensing, the number of available datasets and benchmarks to tackle SITS analysis have increased in quantity over the last decade [69]. Moreover, the characteristics of these datasets were improved over time by increasing the number of labeled data, the number of acquisitions (i.e., the length of the time series) and the proposed modality.

This increase in datasets is also the result of open access to land cover information, especially in agriculture [70]. A notable example is EuroCrop [71], which harmonizes the self-declarations made by farmers in Europe.

The Satellite Image Time Series Datasets GitHub repository [72] lists about 50 SITS datasets, collected between 2017 and 2023. Thirty-one of these have been annotated for classification tasks and one for an extrinsic regression task. The most common SITS source is Sentinel-2, which provides all the data for fourteen datasets and contributes to another thirteen datasets. Sentinel-1 and Landsat-8 are also common sources, used in 13 and 6 datasets, respectively. Spatial resolution of the datasets ranges from 3m to 60m and temporal resolution from daily to monthly. In the rest of this section, we provide a brief description of six of the most used datasets (measured using the number of citations at the current date of publication). Information regarding other datasets is available from the aforementioned GitHub repository and the references provided there.

The Time Series Land Cover Classification Challenge (TiSeLaC) dataset [73] contains a one-year time series of 23 time steps for Reunion Island. The data were collected from the Landsat 8 OLI in 2014, and contains three derived indices, as well as the seven spectral bands. The dataset contains almost 82,000 pixel-level samples covering nine land cover classes. This is the first dataset made available for SITS classification, and it was proposed as a benchmark at the European Conference on Machine Learning conference. Since its introduction, the random train-test split procedure performed at the pixel level has been revised to ensure independence between training and testing data.

The BreizhCrops database [74] contains one-year time series of Sentinel-2 images for Brittany, France for both 2017 and 2018. The data includes two processing levels: The top-of-atmosphere raw reflectance data, and the bottom-of-atmosphere data, which has been corrected so images share the same reflectance scale. The time series are of varying lengths, with either 51 or 102 images for the top-of-atmosphere time series and an average of 53 images for the bottom-of-atmosphere time series. The database is split into datasets covering the four different sub-regions of Brittany allowing easy splitting into spatially separate training, validation, and test datasets. The task is crop type classification and there are nine classes. The data are parcel level, and bands are mean aggregated over each parcel.

Dynamic Earth NET dataset for Harmonized, inter-Operable, analysis-Ready, daily crop monitoring from space (DENETHOR) [75] is a high resolution (3m) daily dataset for crop type classification and change detection. It includes data collected for 2018 and 2019 from Sentinel-1, Sentinel-2, and Planet Fusion for two tiles located in Northern Germany. The time series data is provided at pixel-level, but parcel information is also provided. Thus, the dataset is suitable for pixel-level or parcel-level analysis. Again, there are nine classes.

TimeSen2Crop [76] is a pixel-based dataset of Sentinel-2 time series for the classification of sixteen crop types in Austria. Each time series is one year long and covers the agronomic year for either 2017/2018 or 2018/2019. As cloud-covered images have been excluded from the dataset, the time series vary in length and have irregular time steps. The dataset contains about one million samples and are conveniently organized by Sentinel-2 tile, with one tile reserved for test data and another for validation data.

Panoptic Agricultural Satellite TIme Series (PASTIS) [77] is a dataset annotated for both semantic and panoptic segmentation. It provides labels for eighteen crop types plus a background class and unique parcel instance labels for all non-background pixels. The dataset spans four different regions of France and the images were collected between September 2018 and November 2019. As images with extensive cloud cover have been removed, the time series lengths vary between 38 and 61 and time between acquisitions is uneven. The multimodal SITS data is composed from Sentinel-1 and Sentinel-2 images.

EuroCrops [71] is a reference dataset for crop classification covering 16 European Union countries. The data is collected from national webpages or geoportals and considerable effort has been made to standardize the data, for example converting crop names in national languages to standard crop codes. Currently, only one year per country is available, typically 2021 although as early as 2018 for France. The number of crop classes also varies, from fourteen classes for Croatia to 320 classes for the Netherlands. In addition to the reference data, the curators have made available the TinyEuroCrops dataset that contains Sentinel-2 time series for regions in Austria, Denmark, and Slovenia [78].

RapidAI4EO [79], [80] is a multimodal dataset consisting of daily Planet and monthly Sentinel-2 images. It comprises 500,000 patches of 640m x 640m sampled from European locations. Year-long time series are provided for each patch, which are annotated with multi-class labels based on the CORINE Land Cover [81] classifications. It was created for the European Union's Horizon 2020 (H2020) program to enable the development of land cover products providing higher levels of spatial and temporal detail than are currently available.

Sen4AgriNet [82] is another benchmark dataset created under the H2020 program. It contains annual time series of Sentinel-2 data that span multiple years. The dataset provides data in both pixel and object format and thus is suitable for a variety of SITS applications.

SITS SOFTWARE PACKAGES

To assist with the development of SITS frameworks, there are dedicated libraries that can process satellite images. These libraries include the Sentinel Application Platform (SNAP) (https://step.esa.int/main/toolboxes/snap/) and the Orfeo Tool Box (OTB) (https://www.orfeo-toolbox.org/) and its deep learning module [83]. There are also dedicated frameworks that can automatically learn from SITS, such as iota2 (https://docs.iota2.net/master/) and the R-SITS package [84].

*PREPROCESSING METHODS*

While much satellite data is provided already preprocessed into analysis ready format, there are several other preprocessing steps that may be required before the SITS data is used in a deep learning model. This section discusses some of the commonly used preprocessing methods.

GAP FILLING AND MISSING DATA INTERPOLATION

A common issue when working with SITS is the handling of missing data. Data may be missing or unreliable due to climate factors, such as cloud or rain cover [85], shadows, aerosols [86], or technical problems such as a sensor malfunction or image loss during data transmission [87]. Additionally, when using SITS spanning large geographical regions, acquisitions from different orbital tracks may be obtained on different days [88]. Multi-modal data can add other complexities. If the modalities have different temporal resolutions, it may be necessary to interpolate the time series in one or more of the modalities to align the time steps. Spatial interpolation may also be needed to align the spatial resolutions of multi-modal data.

Generally, simple interpolation methods are adequate. Linear [88], [89] or cubic spline [90], [91] interpolation methods are commonly used for temporal interpolation, while nearest neighbor [86], [91] or bilinear [92] interpolation can be used for spatial interpolation. However, more rigorous methods can be adopted when necessary. Inverse distance weighting assigns a weight to each interpolation point based on its distance from the prediction point, however it assumes independence between the interpolation points [93]. Kriging is a popular geostatistical method that models the spatial relationship between the interpolation points [94], thus accounts for correlations between the points.

The methods discussed so far assume a linear relationship and homogeneity between the missing data and the data used for interpolation [95]. In many tasks, these assumptions may not be valid. In these cases, statistic machine learning methods such as decision trees [95] and Gaussian process-based methods [92] have been used.

Although the above techniques are effective when only a few data points are missing, they may be ineffective if there are large gaps in the time series. For example, in areas with high cloud cover such as the tropics, usable optical images may not be obtainable over periods of several days or weeks [96]. Data fusion techniques combining microwave data (which are not affected by clouds, nor do they require illumination by an external source) with the optical data show potential for addressing this issue [97] by augmenting these images and/or interpolating the missing pixels [98]. Examples include the transformer temporal-spatial model (TTSM) [99] and a generative adversarial model, CycleGAN [100], which both used Sentinel-1 data to gap-fill Sentinel-2 time series.

HANDCRAFTED FEATURES

Spectral indices, which are features computed from two or more spectral bands, are commonly used in place of or to supplement the bands. The most commonly used index is the normalized difference vegetation index (NDVI) [101], which compares the values of the red and near-infrared (NIR) band using this formula:

$$NDVI = \frac{NIR - Red}{NIR + Red}. \tag{1}$$

High NDVI indicates high levels of vegetation, as the red frequencies are absorbed by vegetation during photosynthesis and NIR frequencies are highly reflected due to high internal leaf scattering [14]. Chaves et al. [14] describe many other possible indices, including the enhanced vegetation index (EVI) [102], and normalized difference water index (NDWI) [103].

Tasseled cap transformations extract principal components from weighted combination of multiple spectral bands [104]. The transformations are designed to provide associations with the biophysical properties of "brightness", "greenness", and "wetness". Additionally, a tasseled cap transformation compresses the spectral bands to a smaller set of independent variables, with minimal information loss. Parameters for tasseled cap transformations are sensor dependent and have for example, been derived for the Landsat OLI [104], MODIS [105], and RapidEye [106]. Moreover, the parameters are usually calculated for vegetative conditions and if used in non-vegetative regions, such as deserts, need to be recalculated for these conditions [107].

Spectral bands and vegetation indices can be used to estimate biophysical variables such as Leaf Area Index (LAI), Fraction of Absorbed Photosynthetically Active Radiation (FaPAR), and Gross Primary Productivity (GPP). These variables have been incorporated into analysis-ready remote sensing products such as MODIS/Terra+Aqua Leaf Area Index/FPAR product (MCD15A3H) [108] and hence can also be used in SITS analysis [109]. Alternatively, they can be derived in the data preparation step for inclusion in DL models [110].

While the use of indices and other derived variables is common when non-DL methods are used, and they continue to be used in deep learning models, some authors have found deep learning methods using the raw spectral bands to be more effective [89], suggesting these methods will learn equivalent information to the index when it is relevant to the task at hand.

CREATING TRAINING AND TEST DATASETS

As is standard practice in machine learning, disjoint datasets are used for training and testing models. The purpose of this data separation is to help ensure reported accuracy is not biased and thus is a reliable measure of the model performance on unseen datasets [111]. Typically, the training data is further split to provide a separate validation dataset that is used to tune model hyperparameters. Cross validation is an alternative method of tuning model hyperparameters. In the k-fold cross-validation approach the training data is divided into $k$ independent partitions, and $k$ models are recursively trained using each of the $k$ partitions in turn to evaluate a model trained on the remaining $k - 1$ partitions. Hyperparameters are set using the configuration giving the highest average performance across the $k$ models. In each case, once the hyperparameters have been selected, the training and validation datasets are combined and used to train the final model. This final model is evaluated using the testing dataset. In deep-learning frameworks, the validation set is often used to avoid overfitting by applying early stopping, causing model training to end at the point where the validation loss starts to increase.

Stratified random sampling may be used to ensure the distribution of a key variable is approximately the same across the datasets. For classification problems, the data is frequently stratified using the class variable [112], [113], while in regression problems stratification may use another key categorical variable. For example, the sampling used for LFMC prediction in [114] was stratified by land cover to ensure the model could predict LFMC across all land cover types.

Another consideration for remote sensing applications is the need for spatial and/or temporal separation between the three datasets. A model that is intended to generalize to other regions needs to ensure the test data comes from regions distinct from the regions in the training data [115]. Similarly, for a model that needs to generalize to other time periods, the test data should represent different time periods than the training data [110]. In some cases, the splitting methodology needs to ensure both spatial and temporal separation between the datasets [116].

NORMALIZING SITS

A common practice in machine learning is to normalize the input features using z-normalization (subtract the mean, then divide by the standard deviation) or min-max normalization (subtract the minimum, then divide by the range) [117]. This ensures that features are all represented at a similar scale, so all features are weighted evenly. In time series classification, rather than normalizing across features at each time step, each individual time series is z-normalized. This allows comparison between time series that have similar trends, but different scaling [118]. However, this leads to losing information about the relative magnitude of the time series, which is important for many remote sensing tasks [119]. Therefore, SITS data is usually normalized per band or feature across all pixels or images, which preserves the temporal profile. Commonly, min-max normalization is used [120]. However, this is sensitive to extreme values. Therefore, Pelletier et al. [119] proposed using the 2% and 98% percentiles instead of the minimum and maximum values, which has since been adopted by many other studies [114], [121]–[123].

## SITS APPLICATIONS

Numerous studies have proposed using SITS with deep learning methods to estimate a wide range of variables. This section presents an overview of a few of the main applications and discusses the reasons for and benefits of using SITS analysis. Related applications are grouped together, showing for instance, the variety of agricultural variables, both categorical and continuous, that can be estimated using SITS analysis and to enable readers interested in a specific field to easily find the relevant detail.

### LAND COVER AND LAND USE

One of the most widely studied applications of SITS is land cover mapping [1], [11], [14]. The objective of land cover mapping is to identify the type of vegetative (e.g., forest, shrubland, grassland, or agricultural uses) or non-vegetative (e.g., bare land, water bodies, urban area) land cover at each point, to generate maps that show the distinct types of land cover across a region. The classes used can range from broad land cover categories (such as forest, grasslands, urban, and water) [124], [125] through to specific crop types [2], [126]. Other related classification tasks include identifying specific land cover or land use features, such as snow cover [127], sinkholes [128], burnt areas [3], flooded areas [129], [130], roads [131], deforestation [132], and forest understory and litter types [112].

### AGRICULTURAL APPLICATIONS

SITS have a significant role in agricultural applications [140], as the time series can capture both the growth cycles of crops and the variation in land use during the year, such as multiple crop cycles and land usage. For many applications, field-level predictions are more valuable than pixel-level predictions [141], [142], thus accurate maps of field boundaries are required [143], [144]. However, field boundary information from farmers is hard to collate over large areas and remote sensing applications for boundary delineation are commonly used [145]. The use of multitemporal satellite imagery can take advantage of phenological differences between different types of crops to improve the accuracy of boundary delineation [141].

Accurate information about the types of crops grown in a region, along with the extent and expected yield of these crops is important to ensure food security and reduce poverty [17]. Crop classification (or crop type mapping), which requires the identification of the crop growing in the location represented by a pixel or land parcel, is an essential step in acquiring this detailed information, but is laborious and error-prone when techniques such as land surveys and farmer self-reporting are used. Automated solutions using satellite data have been widely studied. Many of the most promising techniques use SITS, as identifying different growth patterns throughout the agricultural growing season is necessary to distinguish between crops. Numerous deep learning solutions have been proposed for crop classification and one of the most promising is the pixel-set encoder and temporal attention encoder (PSE-TAE) [146], which has been adopted in many recent studies [75], [123], [147], [148]. Besides, various applications (e.g., yield forecasting, transport planning, or irrigation management) require predictions before the end of the agricultural season. This task, known as in-season or early crop type mapping in remote sensing, intends to predict as early as possible the crop for each field in a scene. Different works tested the ability of specific architectures to provide accurate predictions at different dates during the year [123], [149]. In the End-to-End Learned Early Classification of Time Series (ELECTS) model, Rußwurm et al. [150] designed a specific loss function that combines a common classification loss with an earliness-rewarded loss, which encourages early predictions when the model is confident.

SITS and deep learning have also been used extensively for crop yield estimation, another agricultural measure important for managing food security [151], [152] and precision agriculture [153], [154]. These models, which are usually designed for a specific crop, aim to estimate the crop yield in tonnes/hectare across an agricultural region. While the studies found predicting crop yield early in the growing season was unreliable, they showed good results for the middle and late parts of the growing season, thus providing advance information about the likely harvest size. Accurate predictions up to two months before harvest have been obtained for crops such as corn, wheat [155], and soybean [156].

Irrigation and water management play an increasing key role in agricultural management both to meet the increasing global demand for food and to meet the challenges of climate change. SITS have been used to detect irrigated and rainfed crops [157]–[159], monitor irrigation extent and frequency [160], and map crop water availability [161].

Regenerative agriculture is used to increase food production through land use diversification. The adoption of integrated crop-livestock systems (ICLS) in Brazil [162] has been shown to provide more efficient use of nutrients and lead to higher crop yields [163]. However, assessing the adoption and effectiveness of ICLS is time consuming, and systems that provide automatic monitoring of ICLS are needed. SITS are useful for these systems as the time series can be used to detect the phases of the ICLS [162]–[165].

Another use of SITS for agricultural applications is to identify abandoned farmland [166]–[168]. Unmanaged abandoned farmland can result in ecological damage such as wildfire risk and soil erosion [167]. However, if these areas are detected and managed, they can be recultivated to increase agricultural production or targeted for ecological restoration such as reforestation [168].

### *SOIL AND VEGETATION MOISTURE*

Soil moisture (SM) is an essential Earth system variable [169]. It plays a significant role in vegetation condition, agriculture, land-atmosphere circulation and influences climatic conditions [170]. Estimating SM from remote sensing data provides an alternative to labor-intensive and time-consuming field measurements [169], [171]. While all of optical, thermal-infrared, and microwave data have been used to estimate SM [171], microwave data is particularly useful. Microwaves can detect the presence of water in soil due to the large difference in the dielectric constant between water and dry soil [172].

Long-term soil moisture products from satellite data are obtainable from the MetOp ASCAT, SMOS MIRAS, and SMAP instruments [173]. However, these have a very low spatial resolution of 40km – 60km. More recently, the ESA Sentinel-1 and German TerraSAR-X and TanDEM-X missions are using SAR technology to provide soil moisture data at spatial resolutions of 10m and 1-2m, respectively [174].

Soil moisture undergoes rapid change, with fast dry-down periods followed by rapid increases from precipitation, thus strategies modelling the temporal behavior of SM are promising for operational retrieval and mapping [175]. SMEN [176] used a combination of SMAP and Sentinel-1 data to achieve a high spatial and temporal resolution soil moisture product for Oklahoma and Tucson in the United States and Geneva, Switzerland.

Soil moisture estimates have also been derived from time series of optical data. These generally have used derived indices and variables, rather than the raw spectral data, such as MODIS normalized difference vegetation index (NDVI) and albedo data [177], or NDVI and normalized difference water index (NDWI) [178].

Combining both the radar and optical data, Efremova [100] used Sentinel-1 and Sentinel-2 data to estimate SM in Australian vineyards, using DL to extract a latent representation and gap-fill missing images in the time series. Zhu [179] proposed two extensions to long-term change detection methods to improve SM estimates in the Yanco agricultural area in Australia from Sentinel-1 and MODIS NDVI data. Muhuri [180] combined RADARSAT-2 time series with a combined Landsat-8 and Sentinel-2 time series for estimating SM in Manitoba, Canada.

Live fuel moisture content (LFMC) measures the water content of living vegetation and is an important indicator of wildfire risk. Ground truth measurements of LFMC are time-consuming to collect as vegetation samples must be collected, weighed, dried, and then reweighed. Therefore, alternative methods of estimating LFMC are needed to provide large-scale information of vegetation conditions.

Deep learning models for estimating LFMC from SITS have received substantial recent attention, as incorporating temporal data allows the models to account for the time-lags between changes to drivers of the vegetation state, such as precipitation [181] and soil moisture [182], and changes to the vegetation state itself, which vary between species [183]. TempCNN-LFMC [114] used year-long time series of MODIS data to estimate LFMC across the contiguous United States (CONUS). Multi-tempCNN [184] extended the TempCNN-LFMC model by adding a time series of meteorological data and creating an ensemble model. Augmenting the optical data with time series of Sentinel-1 SAR data is a promising technique, due to the microwave scattering caused by water in vegetation. Rao et al. [185] combined 3-month time series of Sentinel-1 SAR and Landsat-8 optical reflectance data to estimate LFMC in the western CONUS. A fourth model [186] combined MODIS, Landsat-8 and Sentinel-1 data and showed that using all three modalities created a more accurate model than using any single modality.

### *SOCIO-ECONOMIC INDICATORS*

While most of the applications of SITS are to estimate environmental and agricultural variables, another important application is estimating socio-economic indicators. Optical satellite imagery has been shown to be effective in identifying slums and areas of informal settlement in urban areas [187]–[190], mapping sites associated with slavery, such as brick kilns [191], estimating rural household poverty [192], and predicting urban pollution levels [193].

An alternative source of remote sensing data is using observations of nighttime lights (NTL) data from the U.S Air Force Defense Meteorological Satellite Program – Operational Linescan System (DMSP-OLS) and the Suomi National Polar-orbiting Partnership (Suomi NPP) Visible Infrared Imaging Radiometer Suite (VIIRS) satellite instruments. The level of nighttime lighting in urban areas is recognized as an indicator of economic development [194], and together with population data can be used to estimate poverty levels [195]. Other proxies of socio-economic measures have been developed from NTL imagery [196], including estimates of well-being not captured in measures such as Gross Domestic Product, such as the informal economy and inequality in human development [197].

Estimating socioeconomic activity from NTL data has shown great promise for country level indicators [198] and provides the ability to perform spatial analysis of economic activity such as measuring regional inequalities [199]. However, models using NTL do not perform well in impoverished areas and cannot distinguish between wealthy sparsely populated regions and poor densely populated areas [200]. Techniques combining the NTL data with daytime images that can distinguish roads and other infrastructure are better able to distinguish between these regions [200].

Static images allow estimation of socio-economic activity at a point in time. However, recent research has shown that simply comparing the results from two estimates from static images does not provide a good indication of how poverty levels are changing over time [196], [201]. Therefore, other studies have used SITS of Landsat and night-light data to predict poverty levels and trends. As poverty levels change reasonably slowly [202], the time series used tend to be multi-year sparse time series, with the data processed into yearly or even three-year composite images. Jarry et al. [203] provided a proof-of-concept, showing that spatiotemporal models predicted the evolution of NTL much better than the spatial-only models did. Similarly, Pettersson et al. [204] used a spatiotemporal model to estimate changes in wealth over time to predict which neighborhoods can be expected to escape poverty by 2030.

## DEEP LEARNING ARCHITECTURES FOR SITS

SITS are multi-dimensional data structures, incorporating two spatial dimensions, the temporal dimension, and the channel dimension (the number and type of bands collected by the sensor, which may be complemented by additional hand-crafted features). These complex data structures require specialized processing to adequately handle the relationships between the dimensions. Many general machine learning algorithms assume independence between all variables in the input – for instance ignoring the temporal ordering in a time series [119]. Recent research has shown the importance of developing specialized machine learning algorithms that can make use of these interdependencies, especially in the temporal domain [205].

This review focuses on the group of machine learning methods known as deep learning. Machine learning methods seek to learn a function that maps a set of predictor variables (in our case, the SITS) onto a target variable by estimating the values of the function parameters from the data. The learned function is then used to make predictions of the most likely value of the target variable, when given a set of new values for the predictor variables. A neural network is a machine learning model that consists of a network of simple processing units called nodes. The neurons are arranged into layers, with the outputs from one layer forming the inputs to the next layer (Figure 1a). The first and last layers are the input and output layers, respectively and the other layers are referred to as hidden layers. Deep learning involves the training of neural networks with multiple hidden layers.

The neurons in the hidden and output layers perform a linear transformation of the inputs followed by a non-linear transformation called the *activation function* (Figure 1b). Common activation functions include the sigmoid, hyperbolic tangent (tanh), and rectified linear unit (ReLU) functions. Each node thus implements a function $y = \phi(\boldsymbol{w}^T \boldsymbol{x} + b)$, where $\phi$ is the activation function, $\boldsymbol{x} \in \mathbb{R}^n$ is the input vector of length $n$, $\boldsymbol{w} \in \mathbb{R}^n$ is a weights vector, $b \in \mathbb{R}$ is the bias, and $y \in \mathbb{R}$ is the output value. During model training, the network is optimized using gradient descent [206] or variants such as Adam [207].

The output layer in a regression model consists of a single linear unit with, in most cases, no activation function. In a binary classifier the output layer is also a single neuron, but it uses a sigmoid activation to give a probability for the membership of the positive class. A multiclass classifier has an output layer containing the same number of neurons as classes and uses a softmax function to output a probability for membership of each class.

The most basic structure of a neural network connects each neuron in one layer to all the neurons in the next layer and is often referred to as either a fully connected neural network or a multilayer perceptron (MLP). Other types of neural networks introduce variations of this structure, such as removing some connections, sharing weights between neurons, and/or adding connections between neurons in the same layer.

Deep learning methods have been highly successful in areas such as image classification, natural language processing (NLP), and time series analysis [5]. They have also been successfully applied to EO tasks and have been shown to be more accurate than the more traditional algorithms such as RF and SVM [119], [208]. The rest of this section reviews the main deep learning architectures that have been used to model SITS data. It first looks at the two most used architectures, Recurrent Neural Networks (RNNs) and Convolutional Neural Networks (CNNs), then discusses more advanced methods, for example, those based on attention mechanisms or spatiotemporal graphs.

### RECURRENT NEURAL NETWORKS

RNNs are designed for sequential data and so are particularly suited to modelling the temporal dimension of SITS datasets [208], however training time is dependent on the length of the time series, making RNNs computationally expensive for long time series [209, Ch. 10]. RNNs process each time step sequentially and generate a "hidden state" that is used as input for the next time step, allowing temporal information to propagate forward (Figure 2) and thus influence subsequence time steps. The hidden state is also passed to the next layer of the model, which could be another RNN layer, a layer using a different architecture, or the output layer. In each layer, the recurrent units that process each time step share parameters. In conventional RNNs, the hidden state is calculated using (2) [210]:

$$h_t^i = \tanh\left(W h_{t-1}^i + U h_t^{i-1}\right) \quad (2)$$

where $W$ and $U$ are learned parameters, $h_t^i$ is the hidden state of the $t^{\text{th}}$ time step in the $i^{\text{th}}$ layer, $h_t^0 = x_t$, where $x_t$ is the $t^{\text{th}}$ time step of the input time. To update the RNN weights, the gradient must be propagated back along the time series and these gradient changes aggregated to derive the final weight update for the cell, a procedure called *backpropagation through time (BPTT)* [211]. For long time series the repeated matrix multiplication incurred can result in the gradient reducing to zero (a *vanishing gradient*) or increasing exponentially (an *exploding gradient*) [212], causing the model to collapse.

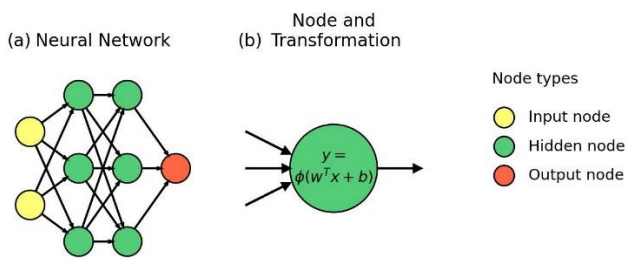

FIGURE 1: Neural Networks and Activation Functions. (a) A fully connected neural network with two hidden layers. (b) Shows the transformation performed by each node. The figure shows a hidden node; output nodes perform a similar transformation.

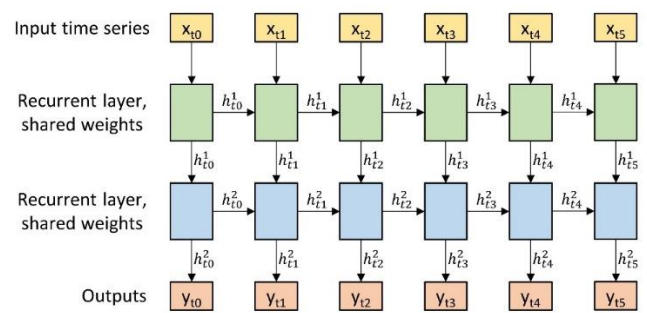

FIGURE 2: An example of a recurrent neural network. At each time step, the recurrent node generates a linear combination of the input for the current time step and the hidden state of the previous time step. This is then passed through the tanh activation function to produce the new hidden state. All recurrent units in a layer share weights, i.e., all green units share weights, as do all blue units.

Several variations of the basic RNN architecture have been proposed to help prevent vanishing and exploding gradients. The most common architectures are long short-term memory (LSTM) [212], gated recurrent unit (GRU) [213], and the more recent stackable recurrent unit (STAR) [214]. These use gates to control the propagation of information along the recurrent layer. Gates are similar to standard neural network nodes and consist of a linear transformation of their inputs followed by either the sigmoid or tanh non-linear function, to produce an output between 0 and 1, or -1 and 1, respectively. Each gate has separate weight matrices, which during training are updated using BPTT.

While Turkoglu et al. [214] showed their STAR network performed well on two remote sensing tasks, classification of the BreizhCrops and TUM datasets, a comparative study of crop type prediction found LSTMs performed slightly better [215]. However, the computational savings from the lightweight nature of STAR units make them attractive for some applications.

RNNs are designed to be unidirectional, so the output for any time step depends on only the input from the previous time step. This is a requirement of tasks such as forecasting, as no future information is available. However, in time series classification the entire time series is available, so information from later in the series may well be useful for extracting features for the current time step. This has led to the development of bi-directional RNNs, which interleave RNN layers that pass hidden states forward through the time series with layers that reverse the flow and pass hidden states backwards [216].

RNNs have been used extensively for modelling SITS, due to their inherent ability to model the temporal context. A list of the studies based RNNs reviewed for this survey is provided in Table 3. RNNs are also frequently used in combination with CNNs and/or attention layers. Further details of these methods are provided in the sections Hybrid Models and Attention and Transformer Models, respectively.

Table 3: Summary of SITS Recurrent Neural Network models.

| Model Name and task | Year and Reference | Architecture |
|---|---|---|
| Land cover classification | 2017 [208] | LSTM |
| Crop type classification | 2017 [217] | LSTM |
| Vegetation quality classification | 2017 [218] | LSTM & GRU |
| Agricultural (crop) classification | 2018 [219] | LSTM & GRU |
| Parcel-based crop classification | 2019 [220] | LSTM |
| Soybean disease detection | 2020 [221] | GRU |
| Atmospheric noise removal | 2021 [222] | Bidirectional GRU |
| Stackable Recurrent cell (STAR) Crop classification | 2021 [214] | STAR |
| Sinkhole detection and classification | 2022 [128] | Bidirectional LSTM |
| Early wildfire detection | 2022 [223] | GRU |
| Live fuel moisture content estimation | 2022 [185] | LSTM |
| Yield mapping and prediction | 2023 [224] | GRU |

One of the first papers to use RNNs for land cover classification was Ienco et al. [208], who showed an LSTM model out-performed non-deep learning methods such as RF and SVM. However, they also showed that the performance of both RF and SVM improves if trained on features extracted by the LSTM model, and in some cases were more accurate than the straight LSTM model.

Rußwurm and Körner [217] compared both an LSTM and the original RNN to two mono-temporal models, an SVM and a 2D-CNN. Both the LSTM and RNN performed significantly better than the mono-temporal models, highlighting the importance of the temporal dimension. They also showed the LSTM model could detect cloud and cloud shadow effects and suggested an LSTM model could reduce the amount of manual data preprocessing needed.

To better exploit cloudy optical SITS, Metzger et al. [225] combined neural ordinary differential equations (NODE) with LSTM and GRU cells. The idea is to use NODE to perform the hidden state prediction while the RNN cell parameters are updated using non-cloudy observations.

The above studies used optical SITS in their models. However, RNN models have also been built using Sentinel-1 SAR data. Ndikumana et al. [219] compared an LSTM and a GRU for agricultural crop classification with classical machine learning models, with both deep learning models showing improved performance over the classical models. Furthermore, they found the GRU model performed slightly better than the LSTM model. In another study, Zhou et al. [220] first up-sampled the SAR data using 2.1 m resolution optical images, then applied an LSTM model to obtain a high-resolution crop classification map.

In other EO applications, Ho Tong Minh et al. [218] compared an LSTM and a GRU to traditional machine learning models (SVM and RF) for vegetation quality classification, showing firstly that the deep learning models achieved better results than the traditional models and secondly, that the GRU performed slightly better than the LSTM model. Rao et al. [185] used an extrinsic regression LSTM model to estimate LFMC in the western CONUS from Sentinel-1 and Landsat-8 data. Kulshrestha et al. [128] identified sinkholes from Sentinel-1A SAR time series using an LSTM model and found using a long time series improved the identification of land deformations leading to sinkholes.

Additionally, GRUs have been used for:

- Disease detection in soybeans [221], where incorporating time series information in the model resulted in a 7% increase in prediction accuracy over non-temporal models.

- Removing atmospheric noise from SAR time series [222]. The deep learning model was able to capture seasonal effects better than traditional methods such as Gaussian filtering.

- Early detection of wildfires [223], where a GRU is used to identify wildfire from a time series from a geostationary satellite. The model identified wildfires earlier than the existing operational products, which is likely due to the extremely high temporal frequency of the time series (15 minutes). Nonetheless, the study shows the ability of GRUs to extract the features needed to identify wildfires.

- Finally, Perich et al. [224] used a GRU for crop yield mapping from sentinel-2 data. This study showed the ability of a GRU model to identify cloud pixels in optical images, thus reducing the need for SITS preprocessing.

### CONVOLUTIONAL NEURAL NETWORKS

Convolutional neural networks (CNNs) are suitable for processing both sequential and multi-dimensional data. They consist of hidden layers of convolutional filters – fixed-sized kernels that slide across the inputs, producing the dot-product between the filter weights and a filter-sized patch of the inputs at each step [226]. This process extracts features comprising contextual information for every point in the input. The filter weights are learned during the training process and the convolutional layer is followed by an activation layer and optionally, a pooling layer (Figure 3). A convolutional filter has the same number of dimensions as the input data (although the size can be one in some dimensions), and convolutions occur in dimensions where the filter size is less than the data size. Convolutional operations are described in terms of the number of dimensions over which they convolve, so a 1-dimensional convolution convolves over a single dimension, irrespective of the number of dimensions in the input data. The output from the convolutional operator has the same dimensions as the convolutional operation. Note (1) if the convolutional filter has a size of one in any dimension, the inputs in that dimension will not be mixed; (2) convolutional operations frequently apply multiple filters in parallel, which provides a multivariate input to the next layer. Table 4 provides a summary of the studies using CNNs reviewed in this paper.

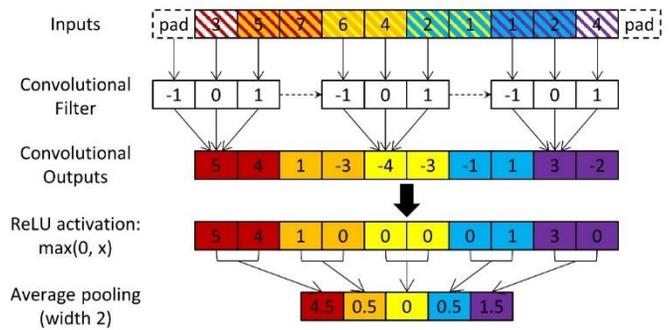

FIGURE 3: An example of a 1D-Convolutional block, including the ReLU activation function and average pooling. The convolutional outputs are created by sliding the convolutional filter along the time series, zero-padding is used to maintain the original time series length. The convolutional filter is applied at every time step; for clarity, the figure shows only a few filter positions. After convolution, the ReLU activation function is applied, followed by average pooling with width two. The colors show which time steps in the input and intermediate time series contribute to each time step in the output.

CNNs have been designed to apply convolutions across the single dimensional temporal domain (1D-CNNs or TempCNNs) [119] and recent studies have shown these TempCNNs outperform RNNs [119], [125]. CNNs have also been designed for the one-dimensional spectral domain [227], and the two-dimensional spatial domain (2D-CNNs) [227]. Three-dimensional CNNs (3D-CNNs) convolve the spatial domain with either the temporal [228] or spectral [229] domain. Several authors have claimed either that CNNs cannot learn temporal features [131], [152], [155], [217], or that RNNs out-perform CNNs for land cover and crop type classification [230]. However, most of these comparisons are to 2D-CNNs, that even when trained on SITS data, ignore the temporal ordering [119]. Other studies have shown using 1D-CNNs to extract temporal information [119], [231] or 3D-CNNs to extract spatiotemporal information [228], [232] are both effective methods of learning from SITS data. The comparison performed on the BreizhCrops benchmark [74] suggests that the performance of 1D-CNNs depends on the preprocessing applied to SITS and might require regularly spaced temporal sampling of the data.

One of the first models to incorporate temporal 1-dimensional convolutions into a land cover classification model was [124], which won the time series land cover classification challenge held as part of the 2017 European Conference on Machine Learning & Principles and Practice of Knowledge Discovery in Databases (ECML PKDD 2017). The success of this model has led to further research into land cover models based on temporal convolutions.

Table 4: Summary of SITS convolutional neural network models.

| Model Name and task | Year and Reference | Architecture |
|---|---|---|
| Land cover classification | 2017 [124] | 1D-CNN & MLP |
| Land cover classification | 2017 [227] | 1D-CNN; 2D-CNN |
| Crop type classification | 2018 [228] | 3D-CNN |
| Temporal CNN (TempCNN); Land cover classification | 2019 [119] | 1D-CNN |
| Summer crop classification | 2019 [231] | 1D-CNN |
| FG-UNET; Land cover classification | 2019 [233] | UNet & 2D-CNN |
| Temporal CNN (TCN) Crop type classification | 2020 [234] | Causal 1D-CNN |
| Eucalyptus plantation identification | 2020 [235], [236] | 2D-CNN |
| Temporal CNN (TempCNN-LFMC); Live fuel moisture content estimation | 2021 [114] | 1D-CNN |
| Temporal CNN (Multi-tempCNN); Live fuel moisture content estimation | 2022 [184] | 1D-CNN |
| 3SI-3D-CNN Crop type classification | 2022 [232] | 3D-CNN |
| Encodeep; Land cover classification | 2023 [237] | 2D-CNN (ResNet-50) |

TempCNN [119] was originally developed as a classification technique for land-cover classification and crop mapping. It consists of three 1D convolutional layers, the output from the final convolutional layer is passed through a fully connected layer, then the final SoftMax classification layer. It was originally used for land cover classification in small regions of France, then for a large-scale classification of the entire state of Victoria, Australia [238]. A similar architecture, using 1D-convolutions, was also proposed in [231] for summer crop identification in the US from univariate Enhanced Vegetation Index (EVI) time series. TempCNN has also been adapted for extrinsic regression [114] and used for LFMC estimation [114], [184], [186].

Another temporal convolutional network, TCN [239], uses *causal convolutions* to ensure that the output from each convolution depends on only the current and previous time steps. Causal convolutions are thus closer in functionality to (unidirectional) RNNs than standard convolutions. In a comparison between TempCNN and TCN, [234] found they performed comparably for crop type classification when evaluated on Sentinel-2 data for Slovenia, but suggested that as

TCN has fewer parameters, it is beneficial when training data or computing resources are limited.

2D-CNNs are mainly used to extract spatial or spatiotemporal features for both pixel and object classification. The model input is usually 4-dimensional — the two spatial dimensions (using a patch centered on the pixel for pixel-based processing), and the spectral and temporal dimensions. The data is convolved spatially, with two main strategies to manage the temporal dimension. In the first strategy, each time step is convolved separately, and the extracted features are merged in later stages of the model [230], [240]. In the second strategy, the time steps and channels are flattened to form a large multivariate image [227], [241]. However, in either method a 2D-CNN used on its own ignores the temporal ordering of the images, so 2D-CNNs are often paired with RNNs or attention blocks to extract temporal features. Further details of these methods are provided in the sections Hybrid Models and Attention and Transformer Models, respectively.

Image-based time series representations [235]–[237] convert a multivariate time series $x \in \mathbb{R}^{n \times d}$, where $n$ is the length of the time series and $d$ its dimensionality, to a two-dimensional representation, a matrix $M \in \mathbb{R}^{n \times n}$. Each element $m_{ij}$ of $M$ is an embedding of the relationship between the $i^{\text{th}}$ and $j^{\text{th}}$ elements of the time series. Popular embedding methods include recurrent plots [242], Gramian Angular Fields (GAF) that account for the temporal correlations, and Markov transition fields [243]. These two-dimensional representations can then be modelled using any computer vision algorithm, such as 2D-CNN or can be used as multiple embeddings, with each resultant matrix forming a separate channel in the input data [236]. A downside of this approach is the increase in complexity, compared to using a 1D-CNN, if the time-series dimensionality is much less than the length of the time series ($d << n$), as is usually the case.

Three-dimensional CNNs (3D-CNNs) can be used to convolve the spatial and temporal dimensions together, combining the strengths of 1D-CNN and 2D-CNNs. In one of the first studies that used a 3D-CNN for SITS classification, Ji et al. [228] found a 3D-CNN crop classification model performed significantly better than the 2D-CNN, again showing the importance of the temporal features. In this study, the 3D-CNN convolved all the optical bands together, so the extracted features contained a combination of the spectral inputs. In contrast, Teimouri et al. [232] designed a 3D-CNN that initially convolves each band separately to fuse optical and SAR images for crop type classification, finding this method performed better than Ji et al.'s method for their chosen task.

### HYBRID MODELS

A draw-back of both RNN and CNN models is that, except for 3D-CNNs, they are unable to process both the temporal and spatial dimensions together. This limits their effectiveness when patch-based per-pixel processing or semantic segmentation is desired. Various architectures that combine recurrent and convolutional layers into one hybrid model have been proposed to overcome this limitation (Table 5). These architectures use the recurrent components to perform the temporal processing, and the convolutional components to perform the spatial processing [244]. Highlighting the importance of these methods, Garnot et al. [230] compared a straight 2D-CNN model (thus ignoring the temporal aspect), a straight GRU model (thus ignoring the spatial aspect) and a combined 2D-CNN and GRU model (thus using both spatial and temporal information) and found the combined model gave the best results, demonstrating that both the spatial and temporal dimensions provide useful information for land cover mapping and crop classification.

Table 5: Summary of SITS Hybrid Models

| Model Name and task | Year and Reference | Architecture |
|---|---|---|
| Land cover classification | 2017 [217] | ConvLSTM & ConvGRU |
| Sequential RNN Crop type classification | 2018 [115] | Bidirectional ConvLSTM & 2D-FCN |
| Crop type classification | 2018 [240] | CNN & GRU |
| Crop type classification | 2019 [245] | 2D-CNN, GRU & attention |
| Crop type classification | 2019 [230] | CNN & GRU |
| Depthwise separable convolution recurrent neural network (DSCRNN); Crop type classification | 2020 [246] | 2D-CNN, LSTM & attention |
| Convolutional LSTM (CLSTM) Crop area estimation | 2020 [247] | 1D-CNN & LSTM |
| Deforestation | 2022 [132] | ConvLSTM & UNet |
| Flood detection | 2020 [129] | Resnet & GRU |
| Road detection | 2020 [131] | UNet & ConvLSTM |
| Spatial-Spectral-Temporal Neural Network (SSTNN) Corn yield prediction | 2021 [155] | 3D-CNN & LSTM |

The first type of hybrid model considered modifies the LSTM or GRU units to replace the matrix element-wise multiplications with convolutions [248], allowing each unit to extract spatiotemporal features. The resulting architectures are referred to as convLSTMs or convGRUs, respectively. In [217], the authors investigated using convLSTM and convGRU for land cover classification from Sentinel-2 SITS and found these models achieved state-of-the-art results with a large number of classes. The results showed that convLSTM and convGRU performed similarly, but the simpler structure of the convGRU units led to faster training of the convGRU model.

Other hybrid CNN-RNN models have separate convolutional and recurrent units, either stacking them sequentially or using them in parallel processing streams [244], fusing the outputs in the final layers of the model. A common approach is to first use CNN layers to extract spatial features from each time step, then process the resulting feature vectors using a GRU [230], [240], [245] or LSTM [131], [155], [246], [247]. However, the high temporal and medium spatial resolution of SITS acquired by low or medium spatial-resolution satellites such as MODIS or Landsat-8 induce a temporal autocorrelation higher than the spatial one [230], leading to the development of "time-first space-later" approaches that make the most the temporal structure in SITS [205]. Hence, for semantic segmentation of Sentinel-2 SITS, Rußwurm and Körner [115] extracted temporal features first, using a bi-directional LSTM, then used a fully convolutional 2D-CNN to incorporate spatial information and classify each pixel in the input patch.

## ATTENTION AND TRANSFORMER MODELS

While RNNs perform well on short sequences, they have several problems when used to learn from long sequences. The back propagation through time process can lead to vanishing gradients, meaning information from early time steps is not propagated to later time steps [249], they have a fixed-size hidden representation that must encode information from all time steps [250], there is no alignment between the input and output representations [251], and the back propagation through time process is inherently sequential and cannot be parallelized [252]. Attention mechanisms, which were first introduced for machine translation, have been proposed for use in conjunction with [250] or in place of [252] RNNs to address these shortcomings. They replace the fixed-size representation with a variable-length memory and align the input representation with the output representations using a soft selection over the input [253], thus providing a more scalable capability.

Attention layers apply a set of attention weights to the hidden representation of a sequence to extract features based on the most important steps in the sequence [5]. For example, in dot product attention [254], the attention weights are constructed from a matrix multiplication of the current hidden state of the model (often called the keys) and a hidden representation of the attention target (a query), followed by the SoftMax normalization function (Figure 4a) [255]. The attention weights are then applied to the values to produce the output representation:

$$A = \text{softmax}(QK^T)V \qquad (3)$$

where $Q \in \mathbb{R}^{n \times d_k}, K \in \mathbb{R}^{n \times d_k}$, and $V \in \mathbb{R}^{n \times d_v}$ represent the queries, keys, and values, respectively. Here, n is the length of the input time series, $d_k$ the dimension of the keys, and $d_v$ the dimension of the values. A commonly used form of attention in SITS DL is self-attention [252], [255], where the queries, keys, and values are three separate linear transformations of the input, so the attention equation becomes:

$$A = \text{softmax}\left(\frac{(XW_q)(XW_k)^T}{\sqrt{d_k}}\right)(XW_v) \qquad (4)$$

where $X$ is the input and $W_q, W_k$, and $W_v$ are the query, key, and vector weights, respectively. Following [252], the attention weights are usually scaled by $\sqrt{d_k}$. Thus, self-attention learns correlations between different steps in a hidden representation of a sequence. In models for multi-dimensional data, self-attention mechanisms can be used across any of the dimensions, resulting in spatial attention [125], channel attention [256] and temporal attention [146], [245] being used in SITS models. In multi-head attention models, several attention units are run in parallel, each unit using different linear transformations of the input for the queries, keys, and vectors (Figure 4b). The outputs from each of the heads are concatenated to obtain the layer output.

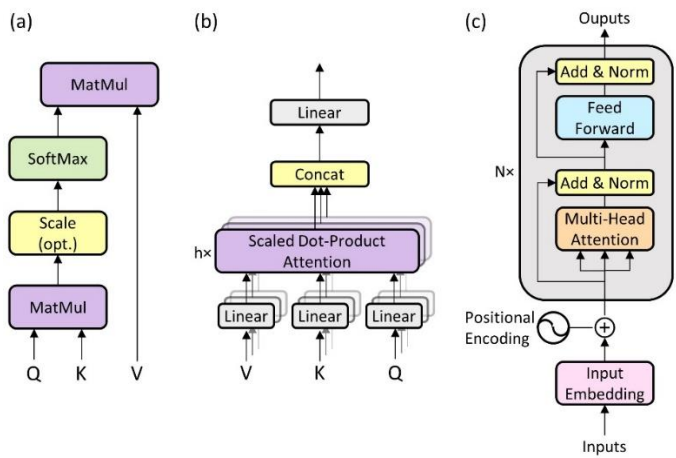

FIGURE 4: Attention Units. (a) A scaled dot product attention unit. (b) A multi-head attention unit; the attention layers are run in parallel. (c) The transformer encoder with positional encoding. Figure re-created from "Attention is all you needed", Figures 1 and 2 by Vaswani et al. [252].

The original use of attention combined RNNs with the attention layers to allow the model to focus on the most important time steps. This type of architecture has been used in several successful SITS models (Table 6). The Object-Based Two-Branch RNN Land Cover Classification (OD2RNN) model [257] used separate GRU layers followed by attention layers to process Sentinel-1 and Sentinel-2 data, combining the features extracted by each source for the final fully connected layers. OD2RNN is an object-based architecture, using the mean pixel values for each time step/band combination as input features. HOb2sRNN [258] refined OD2RNN by using a hierarchy of land cover classifications; the model was pretrained using broad land cover classifications, then further trained using the finer-grained classifications. DCM [259] and HierbiLSTM [260] both use a bi-directional LSTM, processing the time series in both directions, followed by a self-attention transformer for a pixel-level crop-mapping model. All these studies found adding the attention layers improved model performance over a straight GRU or LSTM model.

Although attention was originally proposed to augment RNN-based models, it has also been used to augment CNN models. TASSEL [125] is an object-based classifier that first uses clustering to extract pixel-based components from each object. Each component is convolved temporally using a 1D-CNN, with weight-sharing between the CNNs. In the last step of the model, attention is used to weight the representation of each component. These weighted representations are passed through fully connected layers to produce the final classification. As well as aiding the classification process, the attention scores were used to construct object saliency maps [125], allowing visualization of the important components of each object.

Channel attention-based temporal convolutional network (CA-TCN) [256] uses channel attention to augment causal TCNs with a channel attention unit. The channel attention mechanism is a squeeze-and-excitation network [267], which first compresses each channel timewise (the squeeze operation), then adjusts the resulting channel attention weights via a gating mechanism (the excitation operation). These channel attention weights are then used to scale the input features.

Attention has also been added to hybrid models. Adding temporal attention to a hybrid 2D-CNN and GRU network resulted in a small performance improvement to crop classification models [245]. DuPLO [244] incorporates attention with parallel CNN and hybrid CNN/GRU branches. For the CNN branch, a 5×5 patch centered on the pixel was extracted for each pixel and the bands and time series were stacked together. The patch stacks were passed through three convolutional layers to extract 1024 features. In the CNN/GRU branch, each time step was first convolved to produce 64 features per timestep. These features were then processed by the GRU coupled with an attention mechanism to produce 1024 temporal features, each a weighted combination of the time steps. The final step of the model was to concatenate the 2048 features and classify using a fully connected network. During training, an auxiliary classifier for each component was used to enhance the discriminative power. TWINNS [262] extended DuPLO to a multi-modal model, using time series of both Sentinel-1 (SAR) and Sentinel-2 (Optical) images. Each modality was processed by separate CNN and convGRU with attention models, then the output features from all four models were fused for the final classifier. Again, an auxiliary classifier for each of the four sub-models was employed during training.

A further limitation of RNNs is that their sequential structure prevents parallel processing, meaning they are slow to train when used with long sequences. The transformer architecture [252] (Figure 4c), which completely replaces RNN layers with parallel self-attention layers (multi-head attention), is designed to overcome this limitation, significantly reducing training times. Positional encoding of the attention inputs provides additional information about each input, which is especially useful in cases where input timesteps are not evenly spaced. Most positional encoding schemes are based on the image acquisition date, however Nyborg et al. [268] investigated thermal positional encodings using temperatures rather than acquisition dates for crop-type identification.

Table 6: Summary of SITS Attention-based and Transformer models

| Model Name and task | Year and Reference | Architecture |
|---|---|---|
| *Models with attention layers* | | |
| Attention-based LSTM (A-LSTM); Winter wheat identification | 2019 [261] | Bidirectional LSTM and attention |
| Temporal attention network (TAN); Crop type classification | 2019 [245] | 2D-CNN, GRU and temporal attention |
| DUal view Point deep Learning architecture for time series classificatiOn (DuPLO); Land cover classification | 2019 [244] | 2D-CNN, GRU, temporal attention |
| TWIn Neural Networks for Sentinel data (TWINNS); Land cover classification | 2019 [262] | 2D-CNN, GRU and temporal attention, multi-modal |
| Object-Based two-branch RNN (OD2RNN); Land cover classification | 2019 [257] | GRU, temporal attention, multi-modal |
| DeepCropMapping (DCM); Corn and Soybean mapping | 2020 [259] | LSTM, self-attention |
| ATtentive weAkly Supervised Satellite image time sEries cLassifier (TASSEL); Land cover classification | 2020 [125] | 1D-CNN, self-attention |
| Hierarchical Object based two-Stream Recurrent Neural Network (HOB2sRNN); Land cover classification | 2020 [258] | GRU, temporal attention |
| LSTM multi-task learning (LSTM-MTL); Rice classification | 2022 [263] | Bidirectional LSTM and attention |
| Hierarchical bidirectional LSTM (HierbiLSTM); Crop type classification | 2022 [260] | Bidirectional LSTM, self-attention |
| Crop type classification | 2022 [264] | 2D-CNN, spectral and spatial attention |
| Channel attention-based temporal convolutional network (CA-TCN); Crop type classification | 2022 [256] | Channel attention |
| *Transformer models* | | |
| Crop type classification | 2020 [255] | Transformer |
| Pixel-set encoder, temporal attention encoder (PSE-TAE); Crop type classification | 2020 [146] | Transformer |
| Lightweight temporal attention encoder (L-TAE); Crop type classification | 2020 [265] | Transformer |
| PSE-TAE; Crop type classification | 2021 [123] | Transformer, multi-modal |
| Transformer Temporal-spatial Model (TTSM); Gap-filling NDVI time series | 2022 [99] | Transformer, multi-modal |
| Global-local temporal attention encoder (GL-TAE) | 2023 [266] | Transformer and 1D-CNN |

Most SITS models that use transformers use only the encoder (Figure 4c). Rußwurm and Körner [255] compared a self-attention model with RNN and CNN architectures. They found that this model was more robust to noise than either RNN or CNN, suggesting self-attention is suitable for processing raw, cloud-affected satellite data. PSE-TAE [146] combined a pixel-set encoder (PSE) (to extract a fixed set of features from objects) with a modified transformer called a temporal attention encoder (TAE) for crop mapping. TAE simplifies the computation of the weight attention matrix by learning a single query for each head. They found that the TAE performed better than either a CNN or an RNN. L-TAE [265] replaced the TAE with a light-weight transformer which is both computationally efficient and more accurate than the full TAE. It further simplifies the weight matrix by using a parameter for the primary query (instead of a linear combination) and by using the channel grouping strategy to effectively reduce the number of parameters [269]. Global-local temporal attention encoder

(GL-TAE) [266] combined an L-TAE with a lightweight convolution (LConv) module, using the L-TAE to extract global features and LCONV to extract local features. They found this architecture achieved similar performance to other, larger, deep learning models on the TiSeLaC and TimeSen2Crop datasets. However, it has far fewer parameters than these other models, thus enabling it to process larger volumes of data.

To better exploit the spatiotemporal structure of SITS, Tarasiou et al. [270] adapt Vision Transformers (ViT), which were originally proposed for natural image classification. Their architecture, Temporal-Spatial Vision Transformer (TSViT), is a full-attentional model that first applies a temporal encoder, then a spatial one.

Transformer encoder/decoder models are well-suited to tasks involving reconstructing missing information from time series, due to its ability to uncover long-term dependencies and auto-regressive structure [252]. The Transformer Temporal-spatial Model (TTSM) [99] was designed to reconstruct missing optical information, such as normalized difference vegetation index (NDVI) in optical time series using information extracted from the fused SAR and optical images.

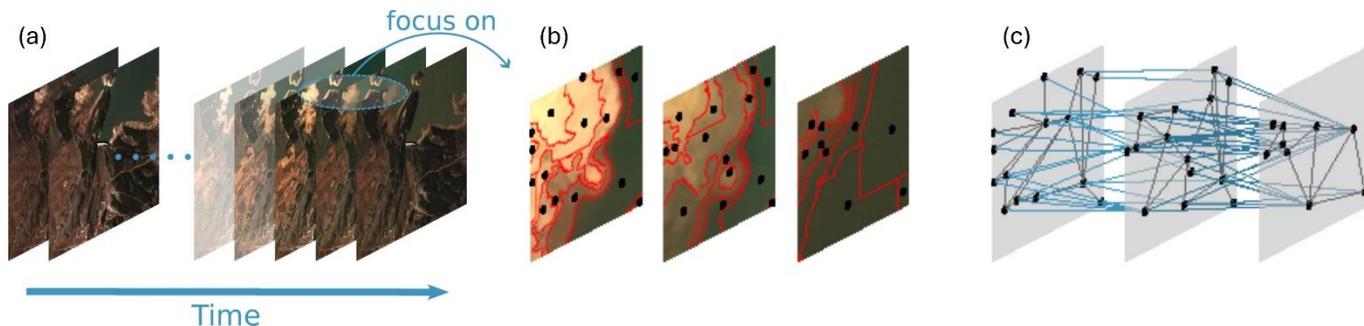

FIGURE 5: A graph created from SITS. The satellite images (a) are first segmented to identify the objects at each timestamp (b). The graph (c) contains a node for each object at each timestamp and two types of edges. Spatial edges (grey) connect adjacent objects at each timestep. Temporal edges connect overlapping objects with adjacent timestamps. Image from "Graph Dynamic Earth Net: Spatiotemporal Graph Benchmark for Satellite Image Time Series", Figure 1 by Dufourg et al. [271]. ©2023 IEEE.

## GRAPH NEURAL NETWORKS

Spatial processing using standard 2D-CNNs assumes the input data is organized in a regular grid format. While this assumption is valid for pixel-based processing, it is invalid for object-based processing. Objects may be irregularly shaped and have more (or fewer) neighboring objects than can be represented by a grid. One method of facilitating spatial learning for object-based modelling is to represent the objects using a graph structure, where the graph nodes represent the objects, and the graph contains an edge between every pair of adjacent objects (Figure 5). Graph neural networks (GNN) methods [272], [273] use this graph representation to learn a network, thus making use of these more accurate spatial and temporal representations of objects [271], [274] and allowing the identification of complex patterns and relationships [275].

According to how the information is passed through the graph, Bacciu et al. [276] distinguish three types of graph neural networks (GNNs). (1) Recurrent GNNs, which assume the state of a node is influenced by the state of its neighboring nodes and so update the node states cyclically until equilibrium. (2) Feedforward GNNs that use multiple layers to form a local context at each step. This type includes convolutional GNNs, which are analogous to standard CNNs, and apply a convolutional operator to each node, aggregating its features with those of its neighbors (Figure 6). However, unlike a standard CNN, the number of nodes involved in each convolutional operation varies. It also includes graph attention networks (GATs) [277] used masked self-attention to learn the relative importance of relationships between nodes and the graph transformer (GT) architecture with a positional encoding based on Laplacian eigenvectors [278]. In the context of spatiotemporal data, spatial-temporal GNNs consider both spatial and temporal dependences by combining graph convolutions to extract spatial features with standard recurrent or convolutional layers to extract temporal features. (3) Constructive GNNs, a special case of feedforward GNNs, perform layer-wise training instead of the more usual end-to-end training approach. This splits the task into smaller sub-tasks, with subsequent layers building on the results from earlier layers.

The use of GNNs for SITS learning tasks has only recently started to be investigated, a summary of these studies is provided in Table 7. One of the first SITS GNN models developed is STEGON [271], a spatial-temporal GNN, which uses one dimensional CNNs to manage the temporal dynamics, then a graph attention mechanism for the spatial information. STEGON performed better than a range of other deep learning models over the two evaluation study areas. Tulczyjew et al. [279] used a graph CNN for semantic segmentation of sentinel-2 images for land cover classification, which compared to U-Net and LSTM models, performed with higher accuracy. Additionally, it was a much smaller model with fewer trainable parameters, thus had both a lower training time and lower memory requirements. Hybrid GCN [280] created two GCN models, one for spatial analysis and one for temporal analysis and demonstrated using the two models together could be used to analyze land cover evolution between 1986 and 2020.

Spatiotemporal graphs model both temporal and spatial dynamics, capturing changes in land cover or land use over time. Spatiotemporal graphs contain edges to connect objects in successive time periods that have overlapping pixels, in addition to the spatial adjacency edges (Figure 5). Once the graph has been constructed, graph deep learning methods are used to train the graph model. Kavran et al. [281] developed a workflow for object-based land cover classification using a CNN for object feature extraction and the GraphSAGE algorithm [282] for node classification. Dufourg et al. [274] paired a spatiotemporal graph with a graph transformer to model land cover classification from the Dynamic Earth Net dataset, showing that learning from an object neighbors' attributes, in addition to its own attributes led to a substantial improvement in accuracy.

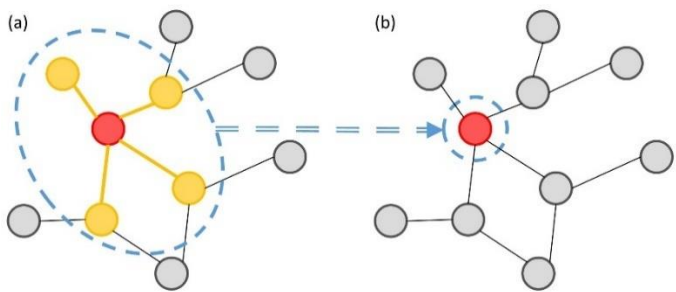

FIGURE 6: Graph Convolution using a neighborhood of one. The orange nodes and edges are convolved with the red node in the input graph (a) to produce the red node in the output graph (b).

Table 7: Summary of SITS Graph Neural Network models

| Model Name and task | Year and Reference | Architecture |
|---|---|---|
| Land cover change detection | 2020 [283] | Spatiotemporal Graph |
| Spatial Temporal Graph Convolutional Neural Network (STEGON) Land cover classification | 2021 [271] | Spatial-temporal GNN |
| Cultivated land map | 2022 [279] | Convolutional GNN |
| Hybrid GCN Land cover change evolution | 2022 [280] | Convolutional GNN |
| Land cover change detection | 2022 [284] | Spatiotemporal Graph |
| Land cover classification | 2023 [281] | Spatiotemporal Graph |
| Land cover classification | 2023 [274] | Spatiotemporal Graph |

Spatiotemporal graphs are well-suited for multi-temporal change detection analysis. In an unsupervised approach, [283] constructed graphs from a time series of bitemporal change maps, and then used graph clustering to analyze the change processes. In [284], graph convolutional layers that incorporate self-attention were applied to a spatial-object temporal adjacency graph to address temporal classification of SITS, for the analysis of temporal evolution of the land cover.

## LEARNING STRATEGIES

The widespread usage of SITS data in Earth observation tasks and applications means a variety of different learning tasks are needed. This section will review some of the main learning tasks used with SITS, including 1) supervised learning for both classification and extrinsic regression, 2) learning from multiple data sources and methods of combining or fusing data from these sources, and 3) learning from less data, including adapting models to new domains.

### SUPERVISED LEARNING

Supervised machine learning is the term given to the process of learning the parameters of a predictive model by training the model using a data set of predictor or explanatory variables and the associated ground-truth values or target variable. The goal is to learn the set of parameter values that, on average, make the best estimates of the labels (the value of the target variable given the predictors) by minimizing a loss function. Once the model has been trained, it can be used to make predictions for the target variable in locations or at times for which we do not have a ground truth measurement.

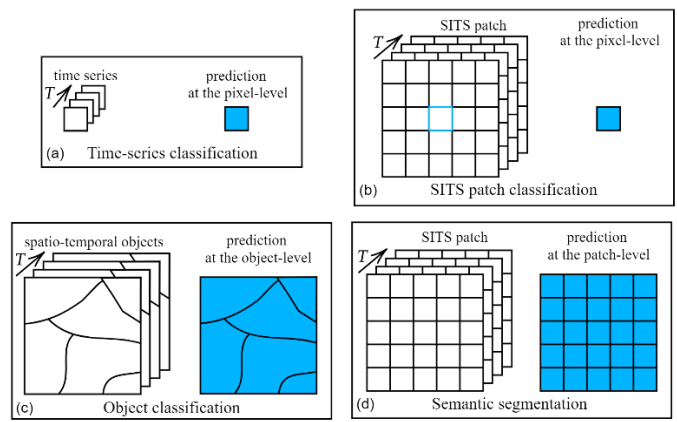

FIGURE 7: Supervised learning for SITS. Top row: per-pixel classification using (a) a time series of pixel data or (b) a time series of patches centered around the pixel being classified. Bottom row: (c) object-based classification first segments the SITS into spatiotemporal objects, which are then classified; (d) semantic segmentation produces a label for each pixel in the patch, allowing segmentation of the patch into regions with the same label.

Supervised learning methods for SITS can also be categorized by the spatial extent of the output variable (Figure 7). Per-pixel methods predict the target value for individual pixels, one at a time [111]. The input may be either a time series of the pixel values, or a time series of patches, centered on the pixel to be predicted. Object-based methods assume the images have been segmented into objects (homogeneous regions), and predict the target value for each object [285]. Semantic segmentation predicts the target value for each pixel in an image patch at once [286]. It incorporates spatial as well as temporal features and allows identification of objects from homogeneous areas within the image.

### PER-PIXEL METHODS

A simple method of analyzing EO data is to train a model to make a prediction for every pixel. Here, a time series is constructed for each pixel by extracting the data for each pixel from each image in the time series. This will usually result in a multi-variate time series where each band (or channel) in the images corresponds to a variable or feature. In some cases, derived features and indices are used to supplement or replace the raw spectral features.

The above method only considers information contained within the pixel itself. However, with the availability of satellite images with moderate and high spatial resolution researchers have hypothesized that information contained in the neighboring pixels may be useful for predicting the pixel value. This had led to the development of spatiotemporal methods. These methods take four-dimensional input, such as a small patch centered on the pixel and the temporal and spectral data for each pixel in the patch. Typically, architectures designed for this method will process the temporal and spatial dimensions in separate layers, for instance by using RNN layers to extract temporal features and CNN layers to extract spatial features [230], [240]. However, studies have also investigated using 3-dimensional CNNs to process all dimensions together [228].

OBJECT-BASED METHODS

At high resolution, pixel classification techniques need to classify many pixels, which is time consuming — for example, producing a global land cover map at 10m resolution requires the classification of about 1.5 trillion pixels (based on a land area of 149M km2 [287]). Furthermore, misclassification of single pixels is common, resulting in a speckled effect on maps produced from the classifier results (often call the "salt-and-pepper effect"). To address both these issues, object-based image analysis techniques can be used [288]. These techniques often lead to better results than the per-pixel methods [289].

Object-based image analysis (OBIA) utilizes a pre-processing step to segment the images into homogeneous areas (called parcels or objects). In crop-type identification tasks, existing vector data identifying the parcels is used to segment the images [290]. In other tasks, a segmentation algorithm is used to identify objects in the image [288]. It is out of the scope of this review to discuss the segmentation algorithms used; reviews of these algorithms are provided in [143], [291]. When segmentation is applied to time series data, care must be taken to ensure homogeneity of the extracted objects over time [292].

Once the time series of images has been segmented, the objects must be transformed to a format suitable for processing by a machine learning algorithm. As objects can contain different numbers of pixels, these methods typically have a feature extraction step, which when applied to SITS, extract a fixed number of features from each time step. These features are then processed by a time series algorithm. Extracted features can be statistical, such as the mean [258] or histograms [293] for each band or index, such as NDVI. However, both clustering (TASSEL, [125]), and neural-network based methods, such as the Pixel-Set Encoder [146] have been used for more complex feature extraction.

While object-based methods reduced the salt-and-pepper misclassifications of per-pixel methods, they do have drawbacks. Errors in the segmentation process can lead to small objects not being detected, thus resulting in areas being misclassified [294]. Two Branches Convolutional Neural Network (TwoBCNN) [294] combined a per-pixel 1D-CNN and an object-based 1D-CNN for land cover mapping. The authors found this technique led to better results than other methods of including spatial information for per-pixel classification.

SEMANTIC AND PANOPTIC SEGMENTATION

Semantic segmentation techniques for SITS take a time series of images and spatially segments it into regions with the same label. The output is a label for each pixel in the SITS. Semantic segmentation extracts both temporal and spatial features and so produces more consistent results than per-pixel techniques (i.e., reduces the salt-and-pepper effect). Compared

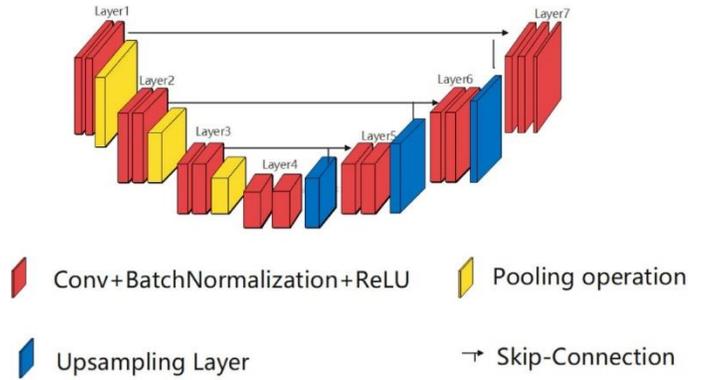

FIGURE 8: A U-Net architecture. Image credit: A Novel Elastomeric UNet for Medical Image Segmentation, Figure 1 by Cai et al. [295]. Used under 

to OBIA methods, it eliminates the need for pre-processing steps to segment the image into objects and extract features from each object.

The main architectures used for semantic segmentation of single images are fully convolutional networks (FCNs) [296], which include U-Nets [297]. FCNs use a series of convolutional and pooling layers to encode the pixel information to a small feature space, then *deconvolution* or up-sampling layers to expand the encoding information back to the required resolution. Two characteristics of U-Nets are (1) the convolution and deconvolution layers are symmetrical (giving the characteristic U shape) and (2) the use *skip connections* that incorporate features extracted in the first convolutional layers directly into the deconvolutional layers (Figure 8).

One of the main challenges with constructing semantic segmentation models for SITS data (as opposed to single images) is the increased model complexity. These models require use of an architecture that can learn from four dimensions, instead of the three dimensions used for single images, while keeping the number of parameters, and consequently the training time, manageable [233]. Semantic segmentation architectures for SITS learning include FG-UNET [233], an architecture for producing land cover maps from time series using ten channels of Sentinel-2 data. FG-UNET handles the temporal dimension by splitting the time series into groups of three steps and concatenating the ten channels of each time stamp to produce images with 30 channels. As there are 33 time steps, each SITS is converted to a set of eleven images. The UNET model processes each of these eleven images individually, then fuses their outputs to make the final predictions. Fully convolutional pixel-wise processing (i.e., a series of $1 \times 1$ spatial convolutions that perform channel mixing for each pixel) is run in parallel with the UNET model to improve separation between classes.

Another fully convolutional model for semantic segmentation is the Pyramid Scene Parsing Network (PSPNet) [298]. PSPNet extracts image features using a ResNet model [299], followed by a layer called the Pyramid Pooling Module (PPM), which has several convolutional layers with different sized filters that are run in parallel. The outputs from these filters are up sampled to the required output size then concatenated. A final convolutional layer produces the segmentation result. Mehra et al. [300] compared PSPNet to several other semantic segmentation methods to produce landcover maps from time series of Sentinel-1 SAR data (they only used the "VV" band). However, rather than use the SITS data as a time series, they first computed statistics for each pixel to provide eleven input channels, which were processed as a single image.

Dual-path interactive network (DPIN) [301] is another dual-path network for the semantic segmentation of Sentinel-2 SITS for paddy rice mapping. The encoder has two processing streams – one processes the time steps separately and the other stacks the time steps. The encoder is multi-staged, and information is exchanged between the two streams between each of the stages. A PPM processes the extracted features and the PPM output processed by the decoder to generate the segmentation map.

Temporal feature-based segmentation (TFBS) [302] is a hybrid LSTM and UNet model for semantic segmentation which was designed for crop mapping using sentinel-1 SAR data. An LSTM is used to learn the temporal features from the SITS data and outputs a single, multichannel image containing these features. This image is then input into a UNet model for the spatial processing and segmentation. The study found that combining LSTM and UNet gave superior results to models based on either one of these architectures.

In another approach that combined LSTMs and UNet (2D U-Net + CLSTM) [6], convolutional LSTMs were embedded in the UNet model. In the encoder layers, each time step was processed independently. The encoded features were then merged using a convolutional LSTM for the decoder layers. This was compared to a 3D UNet (without the LSTM layers), and while both models gave good results, the 2D U-Net + CLSTM model generally performed the best. Tarasiou et al. [303] found they could improve the results of both models by pretraining the models using contrastive learning. Their proposed context-self contrastive loss (CSCL) was especially effective at improving performance at object boundaries, which are often where semantic segmentation performs poorly.

A limitation of semantic segmentation for some applications is that it does not identify individual objects. In crop mapping, if the image contains two wheat fields, all the pixels in these fields will simply be classified as "wheat". If the task requires identification of which pixels belong to each wheat field, then a related method called panoptic segmentation can be used. In panoptic segmentation, each pixel is assigned two labels – one identifying the object (i.e., an object number) and the other identifying the class of the object (e.g., wheat). The difference between the two is shown in the image of an urban street (Figure 9a). Figure 9b shows the semantic segmentation of this image, where all buildings are labeled the same. Figure 9c shows the individual objects that need to be identified, the non-highlighted parts form the background. Figure 9d shows panoptic segmentation, where as well as assigning a class label (such as building), each object is also identified. The background is

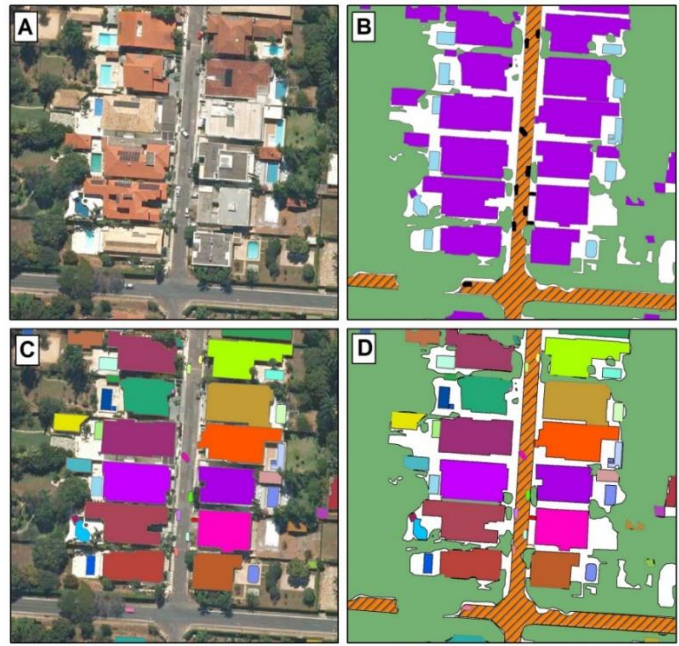

FIGURE 9: Semantic, instance, and panoptic segmentation. Image credit: Panoptic Segmentation Meets Remote Sensing, Figure 1 by de Carvalho et al. (2022) [304]. Used under 

classified (e.g. as road or garden), but not separated into objects [304].

The first work to develop a panoptic segmentation algorithm for SITS consists of two modules [77]. The first module, called U-TAE (UNet with Temporal Attention Encoder) is a modified UNet that incorporates L-TAEs between the layers of the encoder and decoder. U-TAE can be deployed on its own as a semantic segmenter, however, for panoptic segmentation it is used to produce a spatiotemporal encoding. The second module, called Parcels-as-Points (PaPs), predicts an object mask and class label from the spatiotemporal encoding. A more recent work [305] proposed the Exchanger encoder, which uses a collect-update-distribute mechanism to learn a SITS feature representation. This was combined with the Mask2Former [306] image segmentation algorithm to produce the panoptic segmentation.

### LEARNING FROM LESS DATA

Although there are large volumes of SITS data available, annotated or labeled data required for supervised learning is often scarce [307]. Furthermore, it is unevenly distributed and can be difficult to obtain, especially in developing countries [308]. Thus, methods of training models using less labeled data are vital for developing EO applications to help address global problems, such as eliminating hunger and monitoring biodiversity [17].

While this challenge is common to various applications, it is exacerbated for SITS applications, which need to allow for various changes in time and space (e.g., crop rotations induce yearly changes in crop types). This subsection reviews methods of addressing this challenge using the copious amounts of unlabeled SITS data. Semi-supervised and self-supervised learning supplement small quantities of labeled data with unlabeled data [309], while unsupervised learning extracts common features from completely unlabeled datasets [307]. When data are available for only some geographical areas and times, domain adaptation techniques address the problem of adapting a model trained on data from one domain to perform effectively on a different domain where there is not enough ground truth data to train new models [310], while meta-learning can benefit from multiple diverse datasets to perform well on other new unseen datasets. In this section, these various learning paradigms are reviewed in the context of SITS.

SEMI- AND SELF-SUPERVISED LEARNING

To get around the limitations associated with the lack of labeled data, semi-supervised learning and self-supervised learning are some of the most promising artificial intelligence solutions to emerge in the past years. Semi-supervised strategies rely on both labeled and unlabeled data during training, whereas self-supervised methods use only unlabeled data to train a network that outputs a latent representation of the data in a low-dimensional space, which could potentially be used for a various range of downstream tasks. The distinction between both learning paradigms is often fuzzy in the literature, with self-supervised learning seen as a specific case of semi-supervised learning in which a small quantity of labeled data is used in a second step to train a predictor for a final downstream task.

Semi-supervised learning techniques usually rely on self-training to generate pseudo-labels for unlabeled data; co-training and tri-training, which exploit two (or three, respectively) views of the data to extract features, thus benefitting from the complementarity of multi-views of the data; or generative models. In the context of SITS, some works used semi-supervised learning strategies with non-deep learning algorithms [311]–[314]. For example, [314] generated several views from SITS based on the extraction of different spectral (*e.g.*, NDVI or NDWI) and temporal (*e.g.*, statistical) features. Inspired by co-training, they exploit the consistency between predictions made by several classifiers on the different features. Using a deep-learning approach, [315] proposed the use of self-paced learning, a pseudo-labelling strategy, to iteratively refine an LSTM network pre-trained with labeled data covering the same study area but collected over past years using pseudo-labels (i.e., model-generated potential labels) to update the model for other (recent) years. In a similar setting, [316] pre-trained a ConvLSTM on the labeled SITS data and then finetuned the pre-trained model by enforcing consistent predictions between two sub-series extracted randomly from the SITS samples. [317] also used consistency regularization with an LSTM network to predict the boreal forest heights in a regression setting. Finally, in the specific case of positive unlabeled learning, in which labels are available only for a positive well-defined class, [318] first trained a variational auto-encoder (based on recurrent networks) on positive samples, then identified as reliable negative samples the ones with the highest reconstruction error, and then finally trained a binary classifier using both positive and reliable labeled samples.

More recently introduced in the literature, self-supervised learning methods are generally divided into three categories: generative, predictive, and contrastive [319], [320].

Generative methods use generative networks (GANs) or auto-encoders to learn a representation of the data. GANs attempt to generate data from random noise that mimic real data by using adversarial learning, whereas auto-encoders learn a compressed view of the data using an encoder-decoder architecture. To our knowledge, these strategies have not been used for SITS in a self-supervised learning context.

Predictive methods are mainly based on neural networks solving pretext tasks (e.g., completing a close sentence, recovering initial data values from introduced noise or masked data, or predicting a color image from a greyscale image) that are derived from the unlabeled training data. Earth observation tasks are particularly suited for pre-trained models as large quantities of EO data are readily available (for example, about 1.3 TBs are generated every day from each of the MSI on the Sentinel-2 satellites [321]), while labeled data can be difficult to obtain [310], especially in remote locations. SITS-BERT [322] is an adaptation of Bidirectional Encoder Representations from Transformers (BERT) [323] for pixel-based SITS classification. Here, the time steps are projected to a high dimension using dense layers and concatenated with positional encoding information. The encoded data is processed by transformer blocks to produce a time-dependent hidden representation. The output layer processes the hidden representation to produce the required predictions. For the pretext task, random noise is added to the pixels, and the model is trained to identify and remove this noise. The pre-trained model is then further trained for required tasks such as crop type or land cover mapping. Similarly, Cropformer masks part of the time series during the pre-training of the network encoder [324]. The final model is used for crop classification. SITS-Former [325] modifies SITS-BERT for patch classification by using 3D-Conv layers to encode the spatial-spectral information, which is then passed through the temporal attention layers. The pretext task used for SITS-Former is to predict randomly masked pixels. U-BARN [326] combines U-Net and Transformers to compute temporal attention at a full spatial resolution. The model was pre-trained using a BERT-inspired pretext task, which consists of reconstructing masked images from a SITS sample.

The last category of self-supervised learning, contrastive learning, enforces similar latent representations for data that are semantically close (e.g., an image and the same rotated image) and, possibly, dissimilar representations for data away from each other in the latent space. These techniques require the use of data augmentation techniques, which have been well-established for natural image processing but are under-explored for SITS. To overcome this issue, Yuan et al. [148] used optical and SAR time series as naturally augmented views to pre-train a Transformer encoder, whereas Adebayo et al. (2023) considered time series acquired at two different periods over the same area as naturally augmented views of each other. Tarasiou et al. [303] pretrained a U-Net-like network using contrastive learning for semantic segmentation. Their proposed context-self contrastive loss (CSCL), based on attention matrix computations, was especially effective at improving performance at object boundaries, which are often where semantic segmentation performs poorly. Marszalek et al. [327] relied on labeled data from past years to form pairs of samples from the same class.

Recently, contrastive learning has been used to generate high-resolution land cover maps from low-resolution labels. Francis et al. [328] treated the multi-class patch labels provided with the RapidAI4EO dataset as a set of fuzzy labels. They used contrastive learning with a learning mechanism they call "ambiguous loss" to associate each pixel with one of the possible classes for the patch.

UNSUPERVISED LEARNING

Finally, unsupervised learning strategies have been explored when no labels at all can be accessed. The main objective is the extraction of common features (*i.e.*, representations) or the discovery of relationships from completely unlabeled datasets [307]. In this context, deep clustering approaches have been proposed for SITS. They usually learn a representation of the time-series data using an autoencoder based on a temporal network for the backbone (*e.g.*, 1D-CNN) that can be then further clustered using traditional clustering algorithms such as k-Means [307]. Lafabregue et al. [329] introduced pairwise constraints within the deep clustering framework. Their contribution involves integrating two constraints based on expert knowledge into the loss function of the autoencoder: if two time series are identified as similar by the experts they must be grouped within the same cluster (must-link), otherwise they must be separated into distinct clusters (cannot-link). In the context of forest fire detection, Di Martino et al. [330] train autoencoders based on 1D-convolutions, 2D-convolution, or 3D-convolutions on SAR time series using a reference period. The fires are detected by extracting deviating time series covering the same area but a different period.

DOMAIN ADAPTATION

To overcome data scarcity, another possible solution is domain adaptation, which addresses the problem of adapting a model trained on data from one *source* domain to perform effectively on a different, but related, *target* domain where there is no or not enough ground truth data to train new models [310]. Domain adaptation techniques try to mitigate the domain gap by reducing domain shift and target shift, which is caused in remote sensing by several factors including differences in weather and surface conditions.

Approaches tailored for SITS have mainly focused on the unsupervised domain adaptation setting, where no target labeled data is available. Boosted by the large literature on machine learning and computer vision, Martini et al. [331] combined a Transformer model with the well-established domain-adversarial neural network (DANN) [332]. In addition to the encoder and classifier modules, DANN introduces a domain classifier, which is trained to not recognize from which domain the data are coming. The objective is to reduce the dissimilarity between source and target features extracted by the encoder so that the classifier is calibrated on the source data to produce accurate predictions for both domains. Training is facilitated using a gradient reversal layer, which acts as the identity during the forward step, and multiplies the gradient by -1 during the backward step. Another popular approach, coined Phenological-Alignment Network (PAN) [333], used a Siamese network with shared weights based on GRU and self-attention mechanism. The dissimilarity between intermediate source and target features is reduced through the computation of Maximum Mean Discrepancy (MMD). It ensures that phenological features are aligned for the two domains.

However, Lucas et al. [334] noticed that non-deep learning domain adaptation techniques applied in combination with Random Forest or TempCNN for SITS were inefficient in the presence of temporal shifts or changes in class distributions. Nyborg et al. [335] also noticed that the performance of the state-of-the-art unsupervised domain adaptation from computer vision was limited for crop-type identification due to their inability to tackle temporal shifts of the crop phenology. Hence, they proposed TimeMatch to account for the temporal shift issue. It combines a temporal shift estimation with a semi-supervised learning framework inspired by FixMatch [336] to adapt a model to the unlabeled target region. Following this work, Match-And-Deform (MAD) [337] leverages optimal transport with dynamic time warping to perform time series matching and timestamp alignment.

While TimeMatch and MAD tackled the problem of cross-region domain adaptation other studies have focused on temporal domain adaptation. In this context, where SITS are acquired over the same study area but cover two different time periods, Capliez et al. [338] showed that more advanced domain adaptation techniques could be more beneficial for SITS. Hence, the authors designed a spatially aligned domain-adversarial neural network (SpADANN) to adapt a model trained on a particular year to other years over the same geographical areas [339]. SpADANN uses DANN and a pseudo-labelling strategy that exploits semantic spatial invariance. It has been further extended and enhanced in a multi-modal context [340]. Jia et al. [341] proposed domain adaptation for sequential data (DAS) that combines LSTM and attention model to discover temporal patterns from the discriminative period with a cyclic GAN to reduce the domain shift between the discriminative periods. They evaluated their approach for cropland mapping and burnt area detection.

Regarding the semi-supervised domain adaptation setting, Lucas et al. [342] proposed Sourcerer, a regularizer applied to the TempCNN model's weights and enabled by the Bayesian learning theory, to account for the quantity of labeled data available in the target domain. While Sourcerer assumes the availability of some labeled data in the target domain, it does not use the source data during the domain adaptation phase.

This makes it suitable for situations where the source data is no longer available.

Meta-learning

For a few years, meta-learning also appeared as an appealing learning paradigm to develop deep learning models that can perform effectively on new unseen tasks with only a few labeled data points. The idea is to share the common knowledge of different tasks by training model parameters on a set of diverse datasets. It is beneficial when access to a small quantity of labeled data is possible for multiple datasets. One of the most promising meta-learning strategies is Model-Agnostic Meta-Learning (MAML) [343]. MAML seeks an initialization of the model's parameters $\Theta$ that facilitates rapid adaptation to new unseen datasets. The algorithm comprises an inner-loop and outer-loop optimization process. In the inner loop, a batch of tasks is sampled. For each task, the model parameters are initialized with a set of parameters $\Theta$, and updated through gradient descent, where the gradient is calculated with respect to the loss function using labeled training data. This inner loop process iterates for a few steps to adapt the model to the specific task. In the outer loop, the model parameters $\Theta$ are updated using a second-order optimization technique, involving the Hessian matrix, based on the loss functions computed in the inner loop. This process is repeated until convergence is achieved. In remote sensing, Rußwurm et al. [344] are the first ones to explore the potential of MAML for a land cover segmentation task. The same authors also applied MAML to times-series data for global land cover classification from 8-day MODIS time-series data [345]. For crop-type identification from SITS, Tseng et al. [346] introduced MAML to train a model for two binary classification tasks: crop *versus* no-crop and a given crop *versus* all other crops.

*CHANGE DETECTION*

Detecting changes between satellite images obtained for the same location but at different times is used to monitor changes to land cover and land use. For instance, detecting urban growth [133], [134] or deforestation [135]. While many change detection methods simply compare two images, more recent research has shown the benefits of incorporating intermediate images.

Papadomanolaki et al. [133] proposed using short SITS of urban scenes to detect changes between the first and last images. They proposed a multitask learning framework that performs semantic segmentation as well as change detection, using a combined UNet and LSTM architecture. They test their architecture using several urban scene datasets and compare the results obtained using bitemporal images to SITS, finding the SITS clearly outperformed the bitemporal data.

UTRNet [136] is an unsupervised change detection method that seeks to address several problems. They note that methods such as that used by Papadomanolaki et al. [133] ignore the interval between the time steps. They claim that when applied to irregular time series, this can lead to the detection of "pseudochanges" – for instance, detecting seasonal changes in vegetation, rather than permanent changes. They address this by incorporating the image time in the model, thus training it to specifically handle irregular temporal sampling. A preprocessing step generates pseudo-labels for the pixels which are used to select training samples. The selection of training samples is done in a manner that ensures a balance between change/no-change pixels and easy/hard to detect changes. These are then used to train a time-distance guided convolutional recurrent network. Once trained, all pixels in the image are classified to produce the change map.

Saha et al. [137] treated the change detection process as an anomaly detection using a self-supervised approach. They first shuffle the SITS, and the LSTM model is trained to rearrange the images (pixels) into the correct order. The test pixels are likewise shuffled, and the model is used to reorder them. Correctly rearranged pixels are assumed to have no change. Incorrectly ordered pixels are assumed to be anomalous due to changes in the time series.

Li et al. [138] were interested in change detection of buildings, which is generally done using very high resolution imagery. They explored how medium resolution imagery can be used for this task, which is challenging as building details are harder to detect at medium resolution scales. The model extracted feature maps from each image at different scales by using a series of Inception layers [139], then used an LSTM to extract temporal features. These features were then upscaled to the original size and convolved to produce the change map.

The methods discussed above focus only on whether a change has occurred during the time spanned by the SITS or not, but do not attempt to determine when the change occurred. Other research seeks to find out not only if a change has occurred, but also when it occurred. Chen et al. [134] for example, use a multi-step process where in the first step, pseudo change labels are learnt between each pair of images, then contrastive learning and feature tracking are used to remove noise from the change images caused by seasonal fluctuations, and finally change maps are produced showing changes between the initial image and each subsequent image. This method showed better suppression of seasonal noise than alternative methods such as UTRNet.

*DATA FUSION AND MULTI-MODAL LEARNING*

The increasing availability of a range of different sources of remote sensing imagery has led to the design of models that combine imagery from multiple sources using multi-modal learning [347]. This may be to increase image resolution in one or more dimensions, to improve SITS quality, to benefit from the different information that may be present in different types of imagery (such as optical images and SAR images) or to supplement the remote sensing data with other types of data (such as weather, soil moisture, or topographical data). These methods take advantage of the complementary information collected by different types of sensors, thus learning more complete information of the target variable [348]. Besides the need for a specific architecture to handle temporal structure of SITS, multi-modal SITS learning poses specific challenges such as handling time series of different lengths and with possible non-aligned time steps.

The process of combining data from multiple sources is called data fusion. Data fusion may be used in pre-processing tasks such as resolution enhancement [349] and interpolating missing data [100]. Alternatively, it may be done as part of the deep learning process, either at the raw data level, or after feature extraction from the individual modalities, or by combining results derived from each modality.

RESOLUTION ENHANCEMENT

A limitation of optical remote sensing is that maintaining an adequate signal to noise ratio involves making a trade-off between spatial, spectral, and temporal resolution [350]. In other words, if we want a sensor that provides high spatial resolution, we must lower either the spectral or temporal resolution. One use of data fusion is to combine images that have high resolution in different dimensions to produce a fused image with a resolution that matches the highest of the source resolutions in each dimension. An example of this type of data fusion is pan-sharpening, which is used to increase spatial and spectral resolution. Some sensors, such as Landsat-8 OLI, provide a panchromatic band (a black and white image covering a wide spectral range) that has high spatial resolution but low spectral resolution in addition to the multispectral bands, which have lower spatial resolution [351]. Pan-sharpening uses the information in the panchromatic image to segment pixels in the multi-spectral image, resulting in a multi-spectral image at the same spatial resolution as the panchromatic image [352]. In an equivalent manner, spatial and temporal resolution can be enhanced by fusing low spatial/high temporal resolution images with high spatial/low temporal resolution images. For example, daily MODIS images (which have low spatial resolution) can be fused with Landsat-8 images (which have 30m spatial resolution but a 16-day temporal resolution) to produce a daily SITS at 30m resolution [353]. Similarly, Wang and Atkinson [354] proposed fusing Sentinel-2 and Sentinel-3 images to produce a daily time series of Sentinel-2-like images.

Another solution to enhance the resolution of low and medium spatial remote sensing images is super-resolution. This ill-posed inverse problem consists of reconstructing a high-spatial resolution (HR) image from a low-spatial resolution (LR) image. In a multi-image super-resolution (MISR) setting, a sequence of LR images is used to reconstruct a HR image [355]. MISR in remote sensing has been catalyzed by the PROBA-V Kelvin challenge (https://kelvins.esa.int/proba-v-super-resolution/data/) proposed in 2018-2019 by the European Space Agency [349], [356], [357]. Two notable outcomes were proposed to address this challenge: (1) DeepSUM that relies on CNN [356], and (2) HighResNet [349] that consists of an encoder, a recursive module, and a decoder. The encoder outputs latent representations for each LR image, which are then fused pairwise with the recursive module. Finally, the decoder outputs the predicted HR image. Razzak et al. [358] revisited this architecture by including a consistency loss to improve the reconstruction, in the context of building delineation. Recently, the field gained further attraction [359] with the release of new datasets such as Worldstrat [360] and MuS2 [361], and the development of new strategies based on attention mechanisms [362] or Transformers [363].

MULTI-MODAL FUSION

An alternative use of data fusion is to provide machine learning models with extra information to enable them to make better decisions. This may be in the form of a second time series, so providing time series of both optical and SAR data as inputs. Other models have combined SITS with time series of non-remote sensing data, such as gridded weather information [184], [364]. A third method combines SITS data with static (also called auxiliary or ancillary) data, which could be a very high spatial resolution image [365], a hyperspectral image, LiDAR point clouds [241] or topographical data (to name a few).

One of the challenges of designing a model architecture for these multi-modal inputs is deciding where the data fusion takes place. The fusion methods used can be categorizing into three main groups [8], [347]:

1. Raw data fusion. In raw data fusion, fusion is done using the data as acquired. If there is no direct correspondence between pixels in the different modalities, then one or both modalities need to be transformed so this correspondence can be established. Typical transformations include resampling data from one modality to match the spatial resolution of the other or interpolating one time series to match the temporal resolution of the other. Raw data fusion is also known as pixel or sub-pixel level fusion. When deep learning models are used, this is sometimes called early fusion.

2. Feature level fusion. In feature level fusion features are first extracted from each source independently, then the extracted features are fused. Further levels of feature extraction may be performed on the fused features before the final model decision is made. In deep learning, feature level fusion is sometimes called middle or layer-level fusion.

3. Decision level fusion. In decision level fusion, an output or decision is made independently for each modality, then the independent outputs are combined, thus forming an ensemble model. As with other forms of ensembling, the ensembled output is expected to be more accurate than the output from any of the constituent models. In deep learning, this is sometimes called late fusion.

Ofori-Ampofo et al. [123] studied fusion methods for crop type mapping from Sentinel-1 and Sentinel-2 time series data using the PSE-TAE model. They investigated using raw data, feature level, and decision level fusion and found there was no clear "best" method; each method performed best for some classes. However, they noted that for under-represented classes raw or feature level fusion is generally better, while greater gains resulted from decision layer fusion for well-represented classes. The study also found that combining high-level temporal features from both optical and SAR time series is beneficial especially when optical data is limited. These results were further confirmed for a variety of supervised tasks including parcel-based classification, semantic and panoptic segmentation [348].

In examples of raw fusion Addimando et al. [364] fused Sentinel-2 and daily weather data for crop-type mapping by merging each Sentinel-2 time step with the closest two weather observations, effectively down sampling the weather data to match the EO data. Rao et al. [185] fused Landsat-8 and Sentinel-1 data for LFMC estimation by interpolating both sources to a 15-day time step. Xie et al. [186], who also estimated LFMC, interpolated MODIS, Landsat-8, and Sentinel-1 data to monthly time steps. Zhou et al. [220] fused time series of Sentinel-1 data with single high-resolution optical images for crop-type classification, sub-sampling the Sentinel-1 data to match the spatial resolution of the optical images.

Feature level fusion is the most common technique for fusing data in deep learning models. While the fusion can occur after any level of feature extraction, it typically occurs between the feature extraction phase and the fully connected layers, or between the encoder and decoder [366], if an encoder/decoder model is used. Typically, when both modalities are time series, a similar architecture is used to process each modality before feature fusion. However, combining static images with SITS requires each branch to use a different architecture.

The most common modalities used for feature level fusion are Sentinel-1 SAR and Sentinel-2 optical data. These fusion models often deploy parallel architectures for processing each modality. OD2RNN [257] and Hob2sRNN [258] both use parallel GRU-based architectures while Teimouri et al. [232] uses parallel 3D-CNNs. TWINNS [262] uses a 4-branch parallel network, deploying 2D-CNN and convGRU branches for each modality. In addition to the main classifier, many of these techniques [257], [258], [262] employ *auxiliary classifiers* that attempt to classify samples using the features from a single modality. The auxiliary classifiers help ensure discriminative but complementary features are extracted from each modality [257]. Another technique that has been used to improve the quality of the extracted features in multi-modal models is *contrastive learning*, a technique that seeks to learn an embedding that places samples from the same class and different modalities close to each other, while separating samples from different classes [148].

Auxiliary information included in models may be a set of single-valued variables for each sample, or a set of multi-dimensional variables, embedding spatial information, for example. Single-value variables, which are mainly used by per-pixel and object-based methods, can be processed directly by the final fully connected model layers [114], [185] or first encoded using, for example, a small MLP [241]. When using spatial auxiliary information, the static input is often first processed by a 2D-CNN to extract spatial features. The extracted features are then merged with the features extracted from the SITS data, and the fused features processed by the final classifier or regressor. For example, MLDL-Net [152] is a 2D-CNN extrinsic regression model, using CNNs to extract time step features, which are then passed through an LSTM model to extract temporal features. Another 2D-CNN is used to extract soil property features. The features from these layers are fused then passed to fully connected layers that combine the feature sets to predict crop yield. M$^3$Fusion [365] is a multi-modal model combining a time-series of Sentinel-2 images with a single SPOT image. The time series was processed by a GRU with attention, while the single image was processed by a 2D-CNN. MMFVE [241] is an extrinsic regression multi-modal model that combines a time-series with a LiDAR point cloud, using 2D-CNNs for the time series data and an MLP for the point cloud.

The final type of fusion, decision level fusion, is less common than the other fusion types. Studies comparing fusion methods have found decision level fusion does not perform as well as the other methods [123], [148], [348]. Also, it can be computationally expensive, as it requires a separate model for each source. An example of a study using decision level fusion is Rustowicz et al. [367], who fused Planet, Sentinel-1, and Sentinel-2 SITS for crop type semantic segmentation in Africa.

When more than two modalities are used, multiple fusion strategies may be deployed. For example, Lahssini et al. [241] fused LiDAR and Sentinel-2 data at the feature level, but each source was also fused with topographic data using raw data fusion.

## ENHANCED LEARNING METHODS

A broad range of deep learning models have been successfully used to model SITS data. However, in some cases enhancements to the basic learning strategy can be used to improve model performance. This section discusses three commonly used enhancements: combining several models into an ensemble, using a hierarchy of class labels, and learning multiple tasks in parallel.

### ENSEMBLE LEARNING

Ensembling methods such random forests (which trains decision trees on random subsets of data and variables) have been shown to improve predictions over those made by single models. When creating an ensemble model, multiple individual models are trained, and during inference the individual model predictions are aggregated to generate the final prediction. Ensembling models improves performance as the aggregation step reduces model variance [368], thus both compensating for errors made by the individual models and reducing the likelihood of the model overfitting. While ensembles are common in some areas of Earth and environmental sciences (weather forecasts and climate models, for example, are frequently based on ensemble models [369]), there are comparatively few examples of ensembling SITS DL models.

One of the easiest ways to ensemble DL models is to train multiple homogeneous models, that vary only in the random weight initialization [370]. Di Mauro et al. [124] ensembled 100 LULC models with different weight initialization by averaging the softmax predictions. They found this produced a more stable and stronger classifier that outperformed the individual models. Multi-tempCNN [184] is an example of an ensemble of homogeneous models for extrinsic regression, where taking the mean of twenty individual model estimates led to improved estimates of LFMC. The authors suggested that as an additional benefit, the variance of the individual model predictions can be used to obtain a measure of uncertainty of the estimates. TSI [126] also ensembles a set of homogeneous models, but instead of relying on random weight initialization to introduce model diversity, the time series are segmented and models trained on each segment.

Another method of introducing diversity is to ensemble models that use a similar architecture but vary some of the hyperparameters. Kussul et al. [227] compared ensembles of 1D-CNNs and 2D-CNNs models for land cover classification from Landsat-8 and Sentinel-1 SITS. Each model in the ensemble used a different number of filters and neurons in the hidden layers, so finding different feature sets useful for classification.

Other methods create ensembles of heterogeneous models. For instance, Xie et al. [186] ensembled three heterogeneous models — a TCN, an LSTM, and a hybrid TCN-LSTM model — for an extrinsic regression model to estimate LFMC. Furthermore, instead of simply taking the mean of the member model predictions, they used a linear regression meta learner to learn the final predictions from the individual model predictions; a technique called stacking. The authors compared this stacking technique to boosting their TCN-LSTM model, using Adaboost [371] to create a three-member ensemble, finding that stacking a diverse set of models out-performed boosting.

### HIERARCHICAL LEARNING

Land cover and crop classes can be constructed in a hierarchy, with broad categories such as agricultural areas at the top level, to more specific categories such as rice fields or vineyards at the lowest level [81]. Lower-level categories that share a parent category have common characteristics which may be useful for modelling. However, many classification models treat the classes as if they are independent. Hierarchical learning uses the relationships between the broad and specific categories to improve model performance.

The Hierarchical Object based two-Stream Recurrent Neural Network (Hob2sRNN) [258] combined sentinel-1 and -2 time series for an object-based land cover classifier, using two streams of GRUs. The model is trained using successive levels of the landcover hierarchy. It is first trained on the top level, then on the next level and so on until the classification level is reached. While training iteration requires a new output layer that corresponds to the classes at that level, the rest of the model uses the weights learned by the previous layer. Training in this manner encourages extraction of features that reflect knowledge learnt from all levels of the classification hierarchy.

The multi-stage, convolutional STAR network (ms-convSTAR) [372] used sentinel-2 data for crop classification in the Swiss cantons of Zurich and Thurgau. The crop classes are organized into a three-level hierarchy. Then a three-stage convolutional STAR model is constructed – each stage is trained to predict the crop classes at the corresponding level in the crop hierarchy. The trained features are incorporated into the next stage. Then, a final CNN stage combines all three probability maps to refine the lowest-level predictions by attempting to enforce consistency between the predictions at the three levels.

The hierarchical category structure-based convolutional recurrent neural network (HCS-ConvRNN) [373] embeds the land cover hierarchy into the architecture and generates attention scores between each class in each layer (of the hierarchy) and the output features. The model then labels each sample layer by layer, until a leaf node is reached. The study results showed the HCS-ConvRNN model performed slightly better than other deep learning models over the study area of China, thus showing the benefit of including hierarchical information in the model.

### MULTI-TASK LEARNING

Multi-task learning (MTL) takes a set of related, but not identical, learning tasks and aims to improve the learning of each task by leveraging knowledge contained in all the tasks [374]. In the supervised learning context, the set of tasks could be multiple classification tasks, multiple regression tasks, or a mixture of classification and regression tasks.

Applications of MTL in SITS include Chimera [375], who combined classification and regression tasks to estimate forest cover types and forest structure metrics from time series of Landsat-7 images and aerial, climate, and elevation data. They used a recurrent CNN architecture and a small ensemble and found the Chimera model improved results in all tasks compared to non-DL methods.

LSTM-MTL [263] is an MTL method for rice mapping. It split the task into multiple tasks based on the region of the United States (i.e., each state represented in the data is treated as a separate task). The model, which is based on an LSTM and attention architecture, first learnt a set of general temporal features, followed by region-specific layers that learnt spatial features, thus learning both common and region-specific features. The study found a significant benefit using the MTL learning approach over DL methods without MTL, showing the potential for using region based MTL learning over large and diverse regions.

Another application of MTL to an agricultural task uses multi-year time series to predict cotton yield at the sub-field level for each of three years (2001-2003) [376]. The field is gridded, and a yield prediction is made for each cell in the grid. The MTL process allowed tasks to be informed by information extracted from other years thus allowing more accurate yield predictions to be made.

MTL has been used to improve models for agricultural field boundary detection [377], a segmentation task. The three learning tasks were to create an extent mask, boundary mask, and distance mask using a UNet model. The study showed including MTL in the segmentation model improved the UNet performance by about 10%, with an increase in the Intersection over Union metric from 0.57 to 0.62.

## CHALLENGES AND FUTURE OPPORTUNITIES

Deep learning shows enormous potential for estimating EO variables from SITS data, and it is likely that many of the methods developed for fields such as computer vision and time series analysis will be successfully adapted for remote sensing tasks. However, there are several challenges presented by SITS that still need to be addressed before the benefits can be fully recognized [378]. This section briefly discusses some of the main challenges.

High resolution satellite data are large, complex datasets [379]; at 10m resolution, about 1.5 trillion pixels are needed for global land coverage, based on a land area of 149M $km^2$ [287]. Current approaches to estimating variables are computationally intensive and are typically deployed for small regions [380]; only a few studies attempt to estimate variables across entire countries [381]. Estimating variables at continental or global scales requires the development of more efficient techniques [382].

Deep learning models can be expensive to train and use, especially the large models that are needed for complex problems [383]. This is exacerbated for learning from SITS, as models need to cope with up to four dimensions. Designing architectures that require fewer resources to train and run and thus make lower demands on scarce environmental resources will be key to ensuring these models do not end up compounding the very problems they are being designed to help address.

Collecting labeled data in sufficient quantities to train machine learning algorithms is often time consuming and expensive [310]; and so may not be routinely collected in some regions [384]. To overcome this, DA techniques can be used to transfer models from regions with plentiful labeled data to regions with little or no labeled data. While several recent advances in DA have been reviewed in this survey, many of the works discussed are concerned with DA between regions that are close to each other; typically, in the same country or continent. More challenging situations are presented when adapting models to regions in different continents [385], where seasonal changes and differences in vegetation types, agricultural and land management practices, and climate conditions [386] are larger. It is potentially harder to adjust for the much larger domain shifts and there is no guarantee that methods that work well under small shifts will work adequately under these larger shifts. Further work to develop methods that can adapt to different geographic domains [342] are needed to help enable global application of EO models.

The multi-dimensionality of SITS data presents challenges when adapting advances in other areas of machine learning. Other remote sensing tasks, such as those that only consider a single point in time, can benefit quickly from advances in computer vision. The inclusion of the time dimension means that far more work is required to adapt these advances for SITS tasks. One example is the use of pre-trained networks, which are available for many remote sensing tasks [385] but at an early stage for SITS [387].

## CONCLUSIONS

Deep learning from SITS has been shown to be effective for modelling EO variables. It can capture complex relationships between variables, outperforming other machine learning and statistical learning techniques. Many families of deep learning architectures have been used to model SITS, with RNNs and 1D-CNNs used to model temporal data, and more complex architectures such as 3D-CNNs and hybrid CNN and RNN architectures used for spatiotemporal modelling.

This review is designed to be a comprehensive resource of the use of deep learning from SITS. It provides Earth scientists and remote sensing specialists with information about the main deep learning architectures, their strengths and weaknesses and the types of tasks that may benefit from the use of these methods.

## ACKNOWLEDGEMENTS

We thank Corentin Dufourg for his valuable inputs on the use of deep-learning graph-based approaches for SITS.

Funding: This work was supported by an Australian Government Research Training Program (RTP) scholarship, the Australian Research Council under award DP210100072, and by a Monash University Postgraduate Publication Award. The corresponding author is Lynn Miller.

## AUTHOR INFORMATION

**Lynn Miller** (lynn.miller@monash.edu) received her B.Sc. (Hons) in computer science from Canterbury University, Christchurch, New Zealand in 1984. She received her Master of Data Science and Ph.D. in data science and machine learning from Monash University, Melbourne, Australia in 2018 and 2023, respectively.

She is currently a Research Fellow in the department of Data Science and Artificial Intelligence at Monash University. Her research interests include using machine learning to estimate environmental indicators from remote sensing data. She is a member of IEEE.

**Charlotte Pelletier** (charlotte.pelletier@univ-ubs.fr) received her electronics engineering degree from INP ENSEIRB/MATMECA, Bordeaux, France and a Master's degree in signal and image processing from Université Bordeaux, Bordeaux, France in 2014. She received PhD in computer science and data science from Toulouse Université, Toulouse, France in 2017.

She is an Associate Professor in computer science at Univ. Bretagne Sud, Vannes, France. She conducts her research at the Institute for Research in Information Technology and Random Systems (IRISA), Vannes, France. Her research interests include time series analysis, tree-based approaches, and deep learning techniques with applications to Earth observations.

Ass. Prof. Pelletier chairs an ISPRS working group on geospatial temporal data understanding (2022-2024) and co-chairs the IAPR Technical Committee 7 on remote sensing and mapping (2021-2024).

**Geoffrey Webb** (geoff.webb@monash.edu), BA (Hons), computer science, La Trobe University, Melbourne, Victoria, Australia, 1982. PhD, computer science, La Trobe University, Melbourne, Victoria, Australia, 1987.

Prof. Webb is a Professor in the Department of Data Science and Artificial Intelligence at Monash University, Melbourne, Victoria, Australia and previously held faculty positions at Griffith University and Deakin University. His research interests include machine learning, pattern discovery, deep learning, and translational artificial intelligence research in health, engineering, and bioinformatics.

Prof. Webb is a member of the ACM. His awards include Pacific-Asia Conference on Knowledge Discovery and Data Mining Distinguished Research Contributions Award, 2022; IEEE International Conference on Data Mining Best Paper Runner-up Award, 2022; Eureka Prize for Excellence in Data Science, 2017; Australian Computer Society ICT Researcher of the Year Award, 2016; Australasian Artificial Intelligence Distinguished Research Contributions Award, 2016; SIAM International Conference on Data Mining Best Research Paper Award, 2018; SIAM International Conference on Data Mining Best Research Paper Honorable Mention Award, 2015; Australian Research Council Discovery Outstanding Researcher Award, 2014; IEEE ICDM Service Award, 2013. He has been both General Chair and Program Committee Chair of the IEEE International Conference on Data Mining. He is a fellow of IEEE.